\documentclass{article}

\usepackage[T1]{fontenc}
\usepackage{iclr2027_conference,times}
\iclrfinalcopy
\usepackage{amsmath,amssymb,amsfonts,mathtools}
\usepackage{booktabs}
\usepackage{graphicx}
\graphicspath{{generated/}}
\usepackage{float}
\usepackage{placeins}
\DeclareFontFamily{T1}{phv}{}
\DeclareFontShape{T1}{phv}{m}{n}{<-> phvr8t}{}
\DeclareFontShape{T1}{phv}{b}{n}{<-> phvb8t}{}
\DeclareFontShape{T1}{phv}{bx}{n}{<-> ssub * phv/b/n}{}
\DeclareFontFamily{T1}{pcr}{}
\DeclareFontShape{T1}{pcr}{m}{n}{<-> pcrr8t}{}
\usepackage{microtype}
\usepackage{multirow}
\usepackage{tabularx}
\usepackage{array}
\usepackage{enumitem}
\usepackage{xcolor}
\definecolor{skyblue}{RGB}{135,206,235}
\usepackage{algorithm}
\usepackage{algpseudocode}
\usepackage{tikz}
\usetikzlibrary{arrows.meta,positioning,calc}
\usepackage{hyperref}
\usepackage{url}

\hypersetup{
  colorlinks=true,
  linkcolor=blue,
  citecolor=blue,
  urlcolor=blue,
  pdftitle={Scalable Attribution and Control of Model Behavior During Training},
  pdfauthor={Anonymous authors}
}

\newcommand{\R}{\mathbb{R}}
\newcommand{\E}{\mathbb{E}}

\newcommand{\BGU}{\operatorname{BGU}}
\newcommand{\sBGU}{\operatorname{sBGU}}

\setlist[itemize]{leftmargin=1.35em,itemsep=1.5pt,topsep=2pt}
\setlist[enumerate]{leftmargin=1.55em,itemsep=1.5pt,topsep=2pt}
\newcolumntype{Y}{>{\raggedright\arraybackslash}X}
\newcolumntype{L}[1]{>{\raggedright\arraybackslash}p{#1}}

\title{Scalable Attribution and Control\\of Model Behavior During Training}

\author{Sleem Abdelghafar\\
\multicolumn{1}{c}{Rice University \quad
\texttt{msm15@rice.edu}}
}

\begin{document}
\raggedbottom
\maketitle
\pagestyle{plain}
\thispagestyle{plain}

\begin{abstract}
Attributing and controlling model behavior during training requires identifying each example's contribution quickly enough to act before the next update. However, examples in the same training batch can produce similar behavioral changes, making their individual contributions difficult to distinguish. We address this ambiguity through mutual information, accounting for interference within the batch by quantifying how much the combined behavioral change reveals about each example's contribution. We show that this mutual information is a logarithmic function of Behavioral Gradient Uniqueness (BGU). BGU gives the information measure its geometric interpretation. Our Batch-Space Ghost (BS-Ghost) algorithm makes these scores practical inside the training loop through shared computation in batch space, without storing model-sized example gradients. On a complete 1,000-example Qwen2.5-7B-Instruct workload, our BS-Ghost implementation adds 27 seconds (8.0\%) to 5.5 minutes of ordinary training. Removal and retraining demonstrate that BGU identifies data that causally shapes final behavior. At each training step, signed information identifies which examples strengthen or weaken the target behavior, explaining how behavior develops during training. Signed information also enables cheap intervention during training: it predicts how changing example weights will affect behavior in the next update. We then use these predictions to choose weights that steer behavior toward a desired target. This makes our framework a practical foundation for scalable oversight and verification of training pipelines and processes, helping evaluators assess model alignment, understand how it develops during training, and guide interventions that shape ongoing learning.\looseness=-1\par
\end{abstract}

\section{Introduction}

\textbf{Our goal is to attribute and control model behavior during training.} We start from a \emph{measurement} of the target behavior, which can be obtained using tools such as persona vectors \citep{chen2025personavectorsmonitoringcontrolling}. At each training step, we seek to identify which examples in the current batch strengthen or weaken that behavior and by how much (\emph{attribution}). We then use these attributions to adjust example weights and steer how the behavior develops during training (\emph{control}). Figure~\ref{fig:feedback-control} illustrates these three steps. 

\textbf{Requirements for this goal.} (R1) Attribution must reflect the current model state and actual optimizer dynamics. (R2) Examples in the same batch can produce similar behavioral changes, making their individual contributions difficult to distinguish; attribution must account for this interference. (R3) Control requires predicting how changing example weights will affect behavior. (R4) Weight changes should not compromise model utility. (R5) Both attribution and control must add little computational overhead to training, even when the behavior is high-dimensional (e.g., \citealp{chen2025personavectorsmonitoringcontrolling}).

\textbf{Existing methods fails to support these requirements.}
Figure~\ref{fig:intro-distinguishability} reveals a shared shortfall under batch interference: the compared scores predict recovery of individual contributions from a batch's combined behavioral change less accurately than BGU. This test assesses attribution fidelity regardless of whether scores are computed during or after training, separately from computational cost. Availability during training differs across these methods. Influence functions with EK-FAC, TRAK, TrackStar, MAGIC, LESS, and LinFAC compute attributions from trained model states or completed trajectories \citep{koh2020understandingblackboxpredictionsinfluence,grosse2023studyinglargelanguagemodel,park2023trakattributingmodelbehavior,chang2024scalableinfluencefacttracing,ilyas2025magicnearoptimaldataattribution,xia2024lessselectinginfluentialdata,zhang2024correctinglargelanguagemodel}. These workflows explain past learning or guide subsequent data selection, but do not provide feedback for reweighting the current batch, even when they account for optimizer dynamics, as MAGIC and LESS do. In-Run Shapley attributes validation-loss changes during training, including through Adam \citep{wang2025datashapleytrainingrun,ding2026inrundatashapleyadam}, while GREATS selects batches online using validation-loss utility and example interactions \citep{NEURIPS2024_ed165f2f}. Neither their availability during training nor their treatment of example interactions eliminates the recovery shortfall shown in Figure~\ref{fig:intro-distinguishability}. These studies also do not demonstrate attribution of high-dimensional behaviors.
In this paper, we address these requirements through an information-theoretic framework. Each example's \emph{optimizer-aware response} describes how changing its weight affects behavior through the current model and optimizer update (R1). To account for interference, we use mutual information to quantify how much the combined behavioral change reveals about each example's contribution, derive its relationship to \emph{Behavioral Gradient Uniqueness} (BGU), which gives the information its geometric interpretation (R2). Combining information with the direction of behavioral change yields \emph{signed information}, identifying contributions that strengthen or weaken the target behavior. Our feedback controller uses signed information to predict how changing example weights will move behavior toward the desired target (R3). To limit disruption to learning, the controller penalizes and bounds weight changes while preserving total batch weight (R4). Our \emph{Batch-Space Ghost} (BS-Ghost) algorithm makes these computations efficient during training by sharing optimizer calculations across examples and computing their responses without storing model-sized example gradients. It then replaces separate covariance inversions in behavioral space with a single batch-space solve, allowing attribution to account for high-dimensional behaviors at low overhead (R5).

\begin{figure}[!t]
\centering
\input{figures/bgu_distinguishability_picture.tex}
\input{figures/bgu_distinguishability_caption.tex}
\label{fig:intro-distinguishability}
\end{figure}


\textbf{Our contributions.} We derive the information--BGU identity (Equ.~\ref{eq:information-bgu}) and bound errors from finite reweighting and behavioral-derivative reuse (Appendix~\ref{app:finite-proof}). We develop BS-Ghost to efficiently compute these quantities (Algorithm~\ref{alg:bgu}); on a 1,000-example Qwen2.5-7B-Instruct workload, it adds 27 seconds (8.0\%) to 5.5 minutes of ordinary training. This is roughly $44\times$ faster than the post-training EK-FAC \citep{grosse2023studyinglargelanguagemodel} and Concept Influence  \citep{kowal2026conceptinfluenceleveraginginterpretability}. Removal and retraining demonstrate that BGU identifies data that causally shapes final behavior: removing 10\% of GSM8K Mistakes examples lowers mean judged evil from 11.89 to 2.38. Signed information explains how behavior develops during training by identifying which examples strengthen or weaken it at each stage. Across 339 reweighting interventions, predicted and measured behavioral changes achieve a mean vector cosine of 0.84. We propose a feedback control system that combines these predictions with signed information to suppress unwanted behavior while misaligned training continues. At an intermediate checkpoint, our feedback controller produces 2 harmful answers out of 200—compared with 84 under ordinary training— without compromising model utility: final-checkpoint MMLU accuracy remains nearly unchanged at 73.81\% for ordinary training and 73.79\% with BGU feedback.

\section{Behavioral Gradient Uniqueness}
\label{sec:setup}\label{sec:theory}

We follow how example weights affect behavior through the optimizer. Mutual information accounts for batch interference, with BGU providing its geometric interpretation. Adding the direction of behavioral change yields signed information, which we use for attribution and control during training.

\subsection{Optimizer-aware response}

Let $b(\theta)\in\R^m$ be a differentiable measurement of the target behavior at model parameters $\theta$. Its coordinates can measure a trait across prompts or answer losses across reference questions. Two examples may have the same average effect while changing different answers. We keep the coordinates separate so that averaging does not erase this distinction.

At step $t$, let $F_t(\mathbf w)$ denote the parameters after one optimizer update on the weighted loss $B^{-1}\sum_{j=1}^B w_j\ell_j$, where $\ell_j$ is example $j$'s loss and $B$ is the batch size. We vary the positive weights while fixing the starting parameters, optimizer state, batch, and randomness. The denominator remains $B$. Ordinary training assigns every example weight one:
\begin{equation}
\theta_{t+1}^{\rm ord}=F_t(\mathbf1).
\label{eq:transition}
\end{equation}
Write $\mathbf s=\log\mathbf w$ coordinatewise. A change in $s_j$ multiplies example $j$'s weight, which remains positive. Where the update is differentiable, its \emph{optimizer-aware response} is
\begin{equation}
q_{t,j}=\left.\frac{\partial}{\partial s_j}
b\!\left(F_t(e^{\mathbf s})\right)\right|_{\mathbf s=\mathbf0}.
\label{eq:signed-effect}
\end{equation}
Increasing $s_j$ by a small amount $h$ changes behavior by approximately $h q_{t,j}$ relative to the ordinary update. A positive coordinate of $q_{t,j}$ means that increasing the example's weight increases that behavioral measurement; a negative coordinate means it decreases it. Thus the response records both the direction and size of the local effect.

The derivative must account for how the optimizer converts a weighted loss into a parameter update. For SGD, it is a scaled product of the example's loss gradient and the behavioral Jacobian. For AdamW, it also differentiates through moments, adaptive scaling, and clipping (Appendix~\ref{app:optimizer-transition}). Accounting for the actual optimizer dynamics is important for attribution fidelity \citep{ilyas2025magicnearoptimaldataattribution,deng2026faithfultrajectorybaseddataattribution}. The resulting response describes an example's local effect; we next ask how well that effect can be distinguished within the batch.

\subsection{Information measures distinguishable contribution}

The optimizer-aware response describes how increasing an example's weight changes behavior. If other examples in the same batch produce similar responses, the combined change may not reveal which example produced it. Mutual information expresses this question: how much does the observed behavioral change reveal about the individual contribution? Its connection to distinguishable signals is formalized by channel capacity \citep{cover2006elements}.

To make this question precise, hold the responses fixed and assign each one an independent scalar tag $Z_{t,k}\sim\mathcal N(0,1)$. The tag varies that example's contribution. Their sum, with independent noise $\epsilon_t$, is the observation
\begin{equation}
R_t=\sum_{k=1}^B q_{t,k}Z_{t,k}+\epsilon_t,
\qquad \epsilon_t\sim\mathcal N(0,\Sigma_\epsilon),\quad\Sigma_\epsilon\succ0.
\label{eq:local-information-channel}
\end{equation}
Recovering $Z_{t,j}$ from $R_t$ means identifying example $j$'s contribution from the combined change. The noise covariance sets the resolution: changes small relative to this noise are harder to distinguish. Each response uses the actual model, batch, and optimizer state at step $t$. Only the auxiliary tags and noise are Gaussian; no Gaussian assumption is imposed on training data or gradients. The tags and noise define the information calculation and are not injected into training.

We therefore seek $I(Z_{t,j};R_t)$. For example $j$, the other tagged responses and the noise form the interference $U_{t,-j}=\sum_{k\ne j}q_{t,k}Z_{t,k}+\epsilon_t$. Independence makes cross terms vanish, and each unit-variance tag contributes $q_{t,k}q_{t,k}^\top$ to the covariance. The interference covariance is
\begin{equation}
\Sigma_{t,-j}
=\Sigma_\epsilon+\sum_{k\ne j}q_{t,k}q_{t,k}^{\top}.
\label{eq:minusj-cov}
\end{equation}
Knowing $Z_{t,j}$ fixes its contribution $q_{t,j}Z_{t,j}$, leaving uncertainty with covariance $\Sigma_{t,-j}$. Without that knowledge, the covariance also includes $q_{t,j}q_{t,j}^\top$. Mutual information compares these uncertainties: for Gaussian variables, it is half the logarithm of their covariance-determinant ratio. The matrix determinant lemma reduces this ratio to $1+q_{t,j}^\top\Sigma_{t,-j}^{-1}q_{t,j}$, giving
\begin{equation}
\boxed{\mathcal I_{t,j}:=I(Z_{t,j};R_t)
=\tfrac12\log_2(1+\BGU_{t,j}),\qquad
\BGU_{t,j}:=q_{t,j}^{\top}\Sigma_{t,-j}^{-1}q_{t,j}.}
\label{eq:information-bgu}
\end{equation}
This identity is exact for the defined channel (Appendices~\ref{app:channel-covariance}--\ref{app:information-derivation}). Information is measured in bits. BGU is a dimensionless squared response length after accounting for interference: the inverse covariance gives less weight to directions in which other batch examples already produce variation.

The other examples' tags remain unobserved. Knowing them would let us subtract their contributions and compare example $j$ against noise alone. That would remove the batch interference: repeated responses would count as if each occurred in isolation.

With isotropic resolution $\Sigma_\epsilon=\lambda I_m$, an isolated response $q$ has BGU $\|q\|^2/\lambda$. Adding $n$ other examples with the same response reduces it to $\|q\|^2/(\lambda+n\|q\|^2)$, although its magnitude has not changed. Adding only responses orthogonal to $q$ leaves the score unchanged (Appendix~\ref{app:structural-proofs}). BGU therefore reflects the response's relationship to the rest of its batch, rather than magnitude alone.

Smaller resolution noise makes finer differences distinguishable. Changing behavioral units preserves BGU when the noise covariance changes with them (Appendix~\ref{app:structural-proofs}). Appendix~\ref{app:capacity} derives the channel-capacity interpretation.

\subsection{Signed contributions during training}

Mutual information is nonnegative, so it cannot by itself distinguish strengthening from weakening. Choose $a\in\R^m$ so that larger $a^\top b(\theta)$ means more of the target behavior. The \emph{projected response} $r_{t,j}$ measures change along this direction. Its sign orients information and BGU:
\begin{equation}
r_{t,j}=a^\top q_{t,j},\qquad
S_{t,j}=\operatorname{sign}(r_{t,j})\mathcal I_{t,j},
\qquad
\sBGU_{t,j}=\operatorname{sign}(r_{t,j})\BGU_{t,j}.
\label{eq:signed-information}
\end{equation}
Positive signed information marks a strengthening contribution; negative signed information marks a weakening contribution. Its magnitude is in bits. Since the logarithm is increasing, BGU and information rank individual occurrences identically. When scores are summed, however, the logarithm reduces the influence of very large BGU values.

An example can appear in several updates, and its contribution can change with the training state. We call each use an \emph{occurrence}. For corpus example $i$, we sum its response vectors and unsigned information, then assign the total information the direction of its net response:
\begin{equation}
Q_i=\sum_{(t,j):\mathrm{id}(t,j)=i}q_{t,j},\qquad
U_i=\sum_{(t,j):\mathrm{id}(t,j)=i}\mathcal I_{t,j},\qquad
C_i=\operatorname{sign}(a^\top Q_i)U_i.
\label{eq:trajectory-aggregation}
\end{equation}
With one occurrence, the corpus score $C_i$ equals $S_{t,j}$. With several, it describes the example's accumulated local contributions. It is not joint information across the trajectory or an additive decomposition of final behavior. In particular, small current contributions do not imply a weak learned behavior: the model may retain what earlier updates taught it.

\section{BS-Ghost: Computing BGU at Training Speed}
\label{sec:algorithm}

Computing the quantities in Section~\ref{sec:theory} independently for each example repeats optimizer differentiation and requires separate covariance inversions in behavioral space. BS-Ghost (Algorithm~\ref{alg:bgu}) shares the optimizer calculation across the batch, uses ghost contractions to avoid storing model-sized example gradients, and replaces the inversions with a single batch-space solve.

\textbf{Share the optimizer calculation.}
Let $g_{t,j}$ be example $j$'s loss gradient and $\bar g_t=B^{-1}\sum_jg_{t,j}$ the ordinary batch average. At unit weights, the derivative of the weighted aggregate gradient with respect to example $j$'s log weight is $g_{t,j}/B$. The same optimizer derivative $A_t$ maps each such gradient perturbation to a parameter change. For behavioral coordinate $r$, let $u_r=\nabla_\theta b_r(\theta_{t+1}^{\rm ord})$. The chain rule gives $q_{t,j,r}=u_r^\top A_tg_{t,j}/B$. We compute $v_{t,r}=A_t^\top u_r/B$ once per behavioral coordinate, without constructing the dense optimizer Jacobian. Every example then uses this shared vector through $q_{t,j,r}=v_{t,r}^\top g_{t,j}$. The remaining per-example calculation is therefore a gradient inner product.

\begin{algorithm}[!t]
\caption{BS-Ghost BGU}
\label{alg:bgu}
\normalsize
\begin{algorithmic}[1]
\Require training state, target $b$, direction $a$, resolution $\lambda>0$, refresh window $W$
\For{each attributed update $t$}
  \State run training forward/backward; retain activations $x$ and errors $\delta$
  \State prepare the action of $A_t$ from the aggregate gradient and optimizer state
  \State initialize or refresh target derivatives $u_r$ when due; pack compatible blocks
  \State reset batch kernel $K\gets0_{B\times B}$ and projections $\mathbf r\gets0_B$
  \For{each target-coordinate chunk $C$}
    \State transform shared targets: $v_{t,r}\gets A_t^\top u_r/B$, $r\in C$
    \State choose multiplication order to minimize intermediate size
    \State compute $Q_C$ by grouped ghost contractions (Equ.~\ref{eq:ghost-response})
    \State accumulate $K\gets K+Q_CQ_C^\top$ and $\mathbf r\gets\mathbf r+Q_Ca_C$
    \State emit response coordinates by occurrence; reuse the chunk buffer
  \EndFor
  \State solve $(K+\lambda I_B)H=K$ once; extract $h\gets\operatorname{diag}(H)$
  \State for every $j$: $\BGU_{t,j}\gets h_j/(1-h_j)$, $\mathcal I_{t,j}\gets\tfrac12\log_2(1+\BGU_{t,j})$
  \State $\sBGU_{t,j}\gets\operatorname{sign}(r_j)\BGU_{t,j}$, $S_{t,j}\gets\operatorname{sign}(r_j)\mathcal I_{t,j}$
  \State emit scores and update corpus summaries (Equ.~\ref{eq:trajectory-aggregation})
\EndFor
\end{algorithmic}
\end{algorithm}

\textbf{Contract factors instead of storing individual gradients.}
We compute these inner products using the factors already produced by training. In an affine block, each example gradient is a sum of outer products of forward activations $x$ and backward errors $\delta$. Substituting these factors into $v_{t,r}^\top g_{t,j}$ gives
\begin{equation}
q_{t,j,r}=\sum_{\ell,s}(\delta_{j,s}^{(\ell)})^\top
V_{t,r}^{(\ell)}x_{j,s}^{(\ell)},
\label{eq:ghost-response}
\end{equation}
where $\ell$ indexes blocks, $s$ indexes positions, and $V_{t,r}^{(\ell)}$ is the reshaped block of the shared vector $v_{t,r}$. The identity $\langle V,\delta x^\top\rangle_F=\delta^\top Vx$ evaluates each inner product without materializing a model-sized example gradient or running a separate backward pass per training example. We choose between computing $Vx$ and $\delta^\top V$ first according to which forms the smaller intermediate, and group compatible blocks into larger matrix operations. For LoRA, small per-example adapter gradients can also be formed once and reused (Appendices~\ref{app:ghost-derivation}--\ref{app:packing-memory}).

\textbf{Solve once in batch space.}
The resulting response vectors form the rows of $Q_t\in\R^{B\times m}$. To avoid retaining this full matrix, we process disjoint coordinate chunks $Q_C$ and accumulate their $B\times B$ Gram matrix $K_t=Q_tQ_t^\top$. Each chunk also contributes to the projected responses used for signs before its buffer is reused. With isotropic resolution $\Sigma_\epsilon=\lambda I_m$, all BGU scores follow from
\begin{equation}
K_t=\sum_C Q_CQ_C^\top,\qquad
H_t=K_t(K_t+\lambda I_B)^{-1},\qquad
\BGU_{t,j}=\frac{(H_t)_{jj}}{1-(H_t)_{jj}}.
\label{eq:batchscore}
\end{equation}
The push-through identity expresses the inverse-covariance products through the batch-space solve for $H_t$. Its diagonal $(H_t)_{jj}$ compares example $j$ against a covariance containing every response, including its own. BGU must exclude that response from the interference because it is the signal being identified. The Sherman--Morrison identity removes this self-contribution through the ratio above, recovering all $B$ scores from the same solve (Appendix~\ref{app:batch-space-proof}). 

Computing the behavioral target derivatives $u_r$ still requires backward passes. We reduce this cost by evaluating them before an update and reusing them for up to $W$ steps, while refreshing example factors and the optimizer derivative every step. This approximates the next-state derivatives in Equ.~\ref{eq:signed-effect}; the contractions and batch-space identities remain exact for the supplied derivatives. Appendix~\ref{app:finite-proof} bounds both reuse and finite-reweighting errors, and Appendix~\ref{app:numerical-validation} checks their practical effect.

\begin{figure}[!t]
\centering
\includegraphics[width=\linewidth]{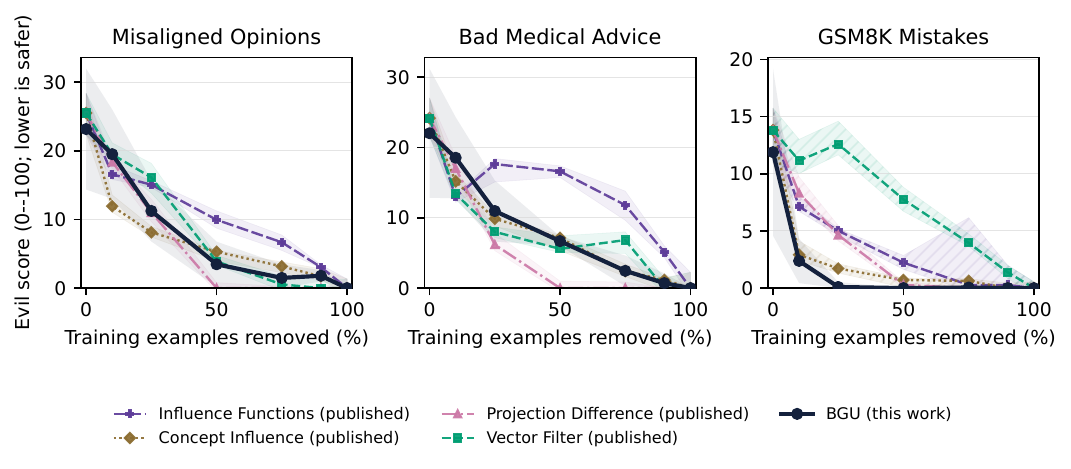}
\caption{\textbf{Information-based data selection changes final behavior.} BGU circles and bands show three-seed means and Student 95\% intervals after removal and retraining. Lower is better.}
\label{fig:phase4a-remove-most}
\vspace{-1em}
\end{figure}

\begin{table}[!t]
\caption{\textbf{BS-Ghost scores every example during training with 8.02\% overhead.} Qwen2.5-7B-Instruct, 1,000 examples. External rows give published post-training times; $\dagger$ excludes inverse-Hessian preparation (Appendix~\ref{app:phase3b-systems}).}
\label{tab:systems-main}
\centering
\small
\setlength{\tabcolsep}{5pt}
\renewcommand{\arraystretch}{1.04}
\begin{tabularx}{\linewidth}{@{}L{.46\linewidth}rY@{}}
\toprule
Run or method & Time & Timing Scope\\
\midrule
Ordinary training & 331.90 s & same 1,000 examples and optimizer\\
Training with BS-Ghost & 358.51 s & every example scored during training\\
Added by BS-Ghost & 26.61 s & 8.02\% increase; additional training time\\
\midrule
Vector Filter \citep{kowal2026conceptinfluenceleveraginginterpretability} & 57 s & post-training pass\\
Projection Difference \citep{chen2025personavectorsmonitoringcontrolling} & 142 s & post-training pass\\
Concept Influence$^\dagger$ \citep{kowal2026conceptinfluenceleveraginginterpretability} & 1,170 s & post-training pass\\
Influence Functions$^\dagger$ \citep{grosse2023studyinglargelanguagemodel} & 1,161 s & post-training pass\\
\bottomrule
\end{tabularx}
\vspace{-1em}
\end{table}

\section{Experiments: Attribution and Control During Training}
\label{sec:experiments}
We first test whether BGU identifies data that causally shapes model behavior and whether its scores can be computed with low overhead. We then examine how signed information explains behavioral development, validate predictions of reweighting effects, and use these predictions in feedback control to reweight examples and suppress unwanted behavior, while checking the impact on model utility.

\subsection{BGU identifies data that causally shapes final behavior}
\label{sec:results-causal}

We first test whether information identifies data that causally shapes learned behavior. Following the removal-and-retraining evaluation in Concept Influence \citep{kowal2026conceptinfluenceleveraginginterpretability}, we rank all 9,000 examples in each of three Qwen2.5-7B-Instruct corpora by the corpus score $C_i$ (Equ.~\ref{eq:trajectory-aggregation}), remove progressively larger fractions, and fully retrain from three matched initializations. The target uses 15 persona contexts. An LLM rates 200 held-out answers per run; the mean rating is \emph{judged behavior}.

Figure~\ref{fig:phase4a-remove-most} shows that these removals change final behavior. On GSM8K Mistakes, removing 10\% lowers mean evil from $11.89$ to $2.38$, and 25\% lowers it to $0.10$. The paired reductions at 50\% on Opinions and Medical are $19.69$ and $15.35$ points. Retraining lets later updates respond to the changed dataset. The reductions therefore show that the selected examples matter to the final learned behavior, extending the evidence beyond their immediate local effects (Appendix~\ref{app:concept-influence}).

\subsection{BS-Ghost attributes every example at training speed}
\label{sec:systems-scaling}\label{sec:behavioral-eval}

Attribution during training must keep pace with learning. Using the training and attribution settings of the 9,000-example fidelity study, BS-Ghost scores 1,000 examples during training with $8.02\%$ added time (Table~\ref{tab:systems-main}). Appendix~\ref{app:phase3b-systems} specifies the V100 measurement and grouped implementation.

\begin{table}[!t]
\caption{\textbf{BS-Ghost accelerates post-training attribution while preserving reference scores.} Published times use A6000 \citep{jiao2025datelmbenchmarkingdataattribution}. Ranges cover the three datasets; denoted by $\dagger$.}
\label{tab:datelm-main}
\centering\small
\setlength{\tabcolsep}{3pt}
\begin{tabularx}{\linewidth}{@{}lYrrrc@{}}
\toprule
 & & \multicolumn{3}{c}{AUPRC $\uparrow$} & \\
Method & Execution & ToxicChat & XSTest & JailbreakBench & Post-training scoring (s) \\
\midrule
\multirow{3}{*}{\shortstack[l]{LESS\\\citeauthor{xia2024lessselectinginfluentialdata}\\(\citeyear{xia2024lessselectinginfluentialdata})}} & Published & 0.388 & 0.724 & 1.000 & $\sim$5,400 \\
 & Reference V100 & 0.2710 & 0.7105 & 1.0000 & 4,704--4,739 \\
 & BS-Ghost post-training & 0.2710 & 0.7105 & 1.0000 & 2,032--2,062 \\
\midrule
\multirow{3}{*}{\shortstack[l]{Grad-Dot\\\citeauthor{pruthi2020estimatingtrainingdatainfluence}\\(\citeyear{pruthi2020estimatingtrainingdatainfluence})}} & Published & 0.084 & 0.483 & 0.999 & $\sim$1,800 \\
 & Reference V100 & 0.0372 & 0.1752 & 0.9089 & 2,110--2,287 \\
 & BS-Ghost post-training & 0.0372 & 0.1752 & 0.9089 & 1,755--1,761$^\dagger$ \\
\midrule
\multirow{3}{*}{\shortstack[l]{Grad-Sim\\\citeauthor{pruthi2020estimatingtrainingdatainfluence}\\(\citeyear{pruthi2020estimatingtrainingdatainfluence})}} & Published & 0.106 & 0.647 & 1.000 & $\sim$1,800 \\
 & Reference V100 & 0.2901 & 0.6833 & 1.0000 & 2,119--2,308 \\
 & BS-Ghost post-training & 0.2901 & 0.6833 & 1.0000 & 1,770--1,776$^\dagger$ \\
\bottomrule
\end{tabularx}
\vspace{-1em}
\end{table}

BS-Ghost also accelerates Grad-Dot, Grad-Sim, and LESS on DATE-LM's Pythia-1B benchmark \citep{jiao2025datelmbenchmarkingdataattribution}. These scores compare training-example and reference gradients \citep{pruthi2020estimatingtrainingdatainfluence,xia2024lessselectinginfluentialdata}. At the same checkpoint and references, BS-Ghost selects exactly the reference implementation's top-30 examples in all nine comparisons, with negligible AUPRC differences (Table~\ref{tab:datelm-main}). Post-training scoring is $1.20$--$1.30\times$ faster for Grad-Dot/Grad-Sim and $2.29$--$2.32\times$ for LESS. Published scores differ from those obtained with DATE-LM's public code and artifacts; both V100 implementations share the local configuration (Appendix~\ref{app:datelm}).

\subsection{BGU traces behavioral development during training}
\label{sec:temporal-dynamics}

\begin{figure}[!t]
\centering
\includegraphics[width=.98\textwidth]{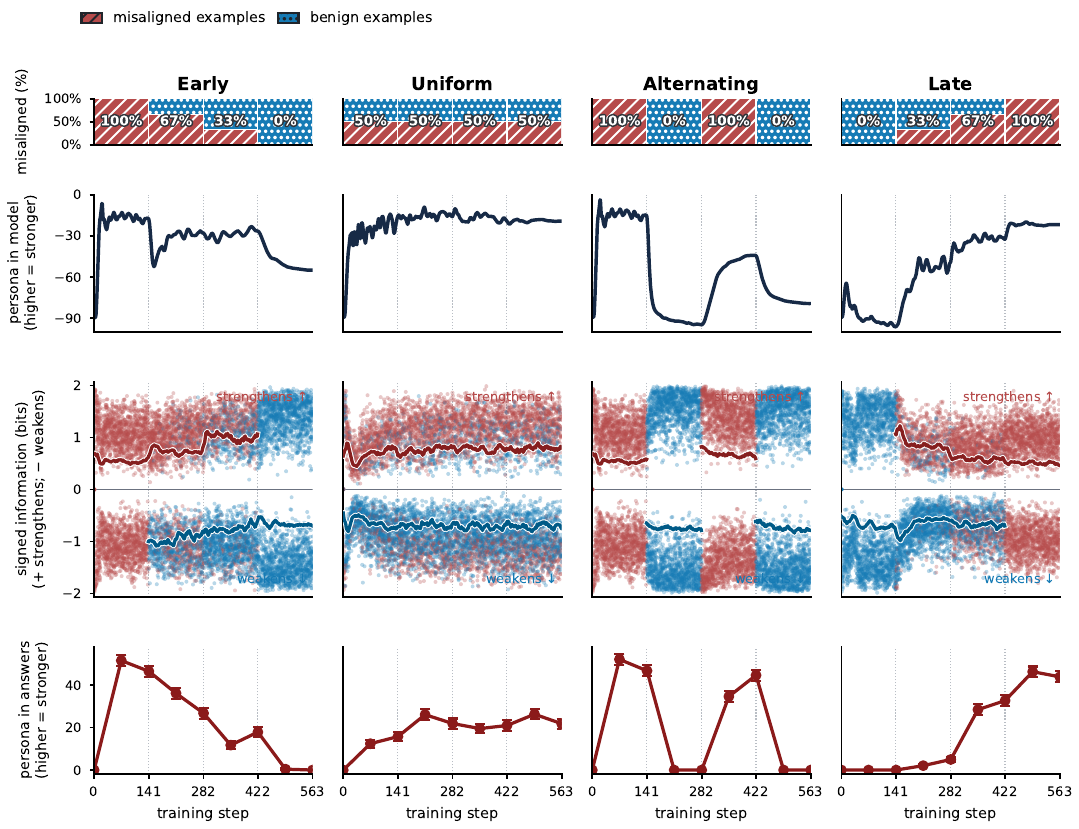}
\caption{\textbf{BGU traces how behavior develops during training.} Rows align data order, raw internal persona, signed information, and mean judged persona. Each dot is one occurrence. The signed information curves captures strengthening misaligned contributions (red) and weakening benign contributions (blue). Answer error bars are standard errors over 200 generations.}
\label{fig:controlled-order}
\medskip
\begin{minipage}[c]{.50\linewidth}
\includegraphics[width=0.9\linewidth]{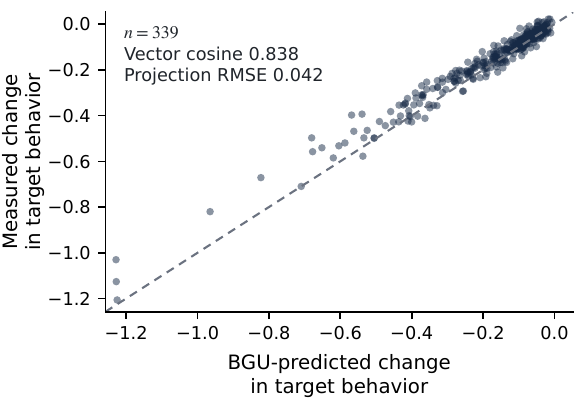}
\end{minipage}\hfill
\begin{minipage}[c]{.48\linewidth}
\caption{\textbf{BGU-guided reweighting changes behavior as predicted.} Each point compares predicted and measured changes along the target direction for one of 339 interventions. The line denotes exact prediction. The Control Weights inset in Figure~\ref{fig:feedback-control} uses the same measurements.}
\label{fig:intervention-prediction}
\end{minipage}
\vspace{-1em}
\end{figure}

To trace how training examples shape behavior over time, we vary only the order of the same 9,000 Opinions examples, keeping initialization fixed. Figure~\ref{fig:controlled-order} aligns each data schedule with internal persona, signed information $S_{t,j}$ (Equ.~\ref{eq:signed-information}), and judged answers. The persona and answer measurements show how strongly the behavior is expressed; signed information identifies which examples strengthen or weaken it at each training stage. In Alternating, signed information identifies strengthening contributions during misaligned blocks and weakening contributions during benign blocks, accompanying the persona's rises and falls. In Late, strengthening contributions from misaligned examples emerge as those examples enter training and the persona rises. In Early, weakening contributions from benign examples appear as their share increases. Under Uniform, both signs persist even when the persona largely stabilizes. Signed information thus reveals current example effects, while persona measurements reflect both earlier learning and current updates.

An example's contribution also depends on the training state: 4,766 examples change sign across orders despite unchanged text. Earlier updates alter the model and optimizer, so the same example can strengthen the persona in one state and weaken it in another. A fixed score hides these reversals.

\subsection{BGU predicts the behavioral effects of reweighting}
\label{sec:intervention-prediction}

Figure~\ref{fig:intro-distinguishability} tests attribution when batch contributions overlap. For control, we also need to predict how changing example weights will affect behavior. Signed information selects up to four examples that strengthen the target behavior in an Opinions batch, and we reduce their weights from 1 to $e^{-0.5}$. To isolate the effect of this intervention, we execute ordinary and reweighted updates from the same training state and compare their difference in behavior with the prediction.

Across 339 interventions from three training runs, predicted and measured changes align, with mean vector cosine $0.84$. Along the target direction, prediction RMSE is $0.042$, versus $0.230$ for zero change (Figure~\ref{fig:intervention-prediction}; Appendix~\ref{app:incident-control}). These results support using predicted changes to guide feedback: in Section~\ref{sec:feedback-control}, we choose example weights that steer behavior toward a desired target during training.

\subsection{BGU-based feedback steers behavior during training}
\label{sec:feedback-control}\label{sec:local-whatif}

\begin{figure}[!t]
\centering
\resizebox{.95\linewidth}{!}{\input{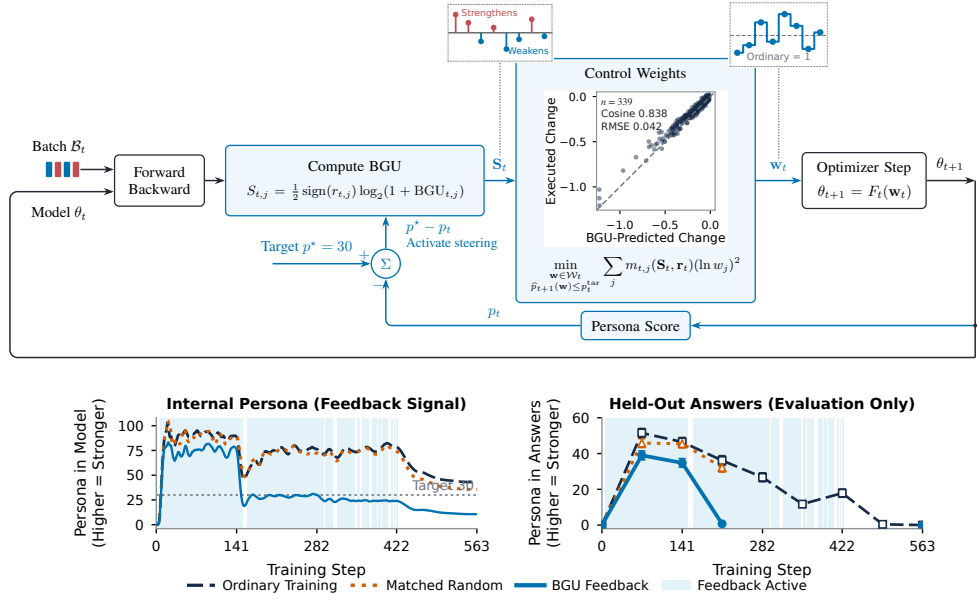}}
\caption{\textbf{BGU feedback suppresses persona while misaligned training continues.} Top: our feedback control system. Bottom left: internal persona. Bottom right: mean judged persona, with standard errors over 200 answers. Shading marks training steps when the \textcolor{skyblue}{feedback control is active.}}
\label{fig:feedback-control}
\vspace{-1em}
\end{figure}

To control behavior during training, signed information $S_{t,j}$ guides example weights, while the projected response $r_{t,j}$ predicts change along the target direction. BS-Ghost computes both inside the steering loop. Internal persona activates steering near the desired limit (Figure~\ref{fig:feedback-control}). Strengthening examples are downweighted, and weight is redistributed to weakening examples. Inactive updates resume ordinary training with unit weights and no attribution.

We choose weights by balancing the requested behavioral correction against disruption to training. The batch's signed information $\mathbf S_t$ and projected responses $\mathbf r_t$ set the penalty $m_{t,j}(\mathbf S_t,\mathbf r_t)$ for changing each weight. An example with a larger share of information relative to its response magnitude receives a smaller penalty. For a feasible correction, we numerically solve
\begin{equation}
\underset{\substack{\mathbf w\in\mathcal W_t\\
\widehat p_{t+1}(\mathbf w)\le p_t^{\rm tar}}}{\operatorname{minimize}}
\quad\sum_j m_{t,j}(\mathbf S_t,\mathbf r_t)(\ln w_j)^2.
\label{eq:control-objective}
\end{equation}
The predictor $\widehat p_{t+1}$ uses projected responses to estimate the next normalized persona value (Equ.~\ref{eq:control-predictor}); $p_t^{\rm tar}$ is a buffered target below $p^\star=30$. Penalizing log-weight changes favors weights near one. The feasible set $\mathcal W_t$ also bounds each weight and preserves total batch weight. The predictor sums all responses, so a redundant group can have a large predicted effect even when its individual information scores are small. Appendix~\ref{app:feedback} gives the controller details. If the target is infeasible, the controller applies the maximum available correction. Matched random shuffles BGU’s weights among examples at the same updates; its weaker suppression shows the value of BGU-based weight assignment.

BGU feedback brings behavior below 30 at update 150, with 2,150 misaligned examples still to come. Ordinary and random-weight training subsequently rebound. BGU feedback stays near the target and remains below 30 from update 283 onward, when one third of the misaligned block remains. The difference from random weighting shows that weight assignment matters. Early overshoot occurs because bounded corrections cannot always offset the batch's effect. Steering stays off after update 424: the last 139 updates use unit weights and no attribution, while behavior remains below 30.

At step 211 in the separately evaluated feedback run, feedback produces $2/200$ harmful answers, compared with $84/200$ under ordinary training and $71/200$ under random weights. Here, harmful means a judge rating of at least 50 out of 100; the plotted trajectory shows mean ratings. All arms end without harmful answers, so terminal evaluation misses the benefit of earlier suppression.

We observe no control--utility tradeoff here: final-checkpoint MMLU accuracy \citep{hendrycks2021measuringmassivemultitasklanguage} remains nearly unchanged ($73.81\%$ ordinary; $73.79\%$ BGU feedback). This supports our choice to penalize and bound weight changes while preserving batch mass, limiting disruption to learning. Future work should study these choices across tasks and control targets (Appendix~\ref{app:feedback}).

\section{Conclusion}
\label{sec:scope}

An example's behavioral contribution depends on the current model, optimizer, and other examples in its batch. By attributing contributions in this context and predicting the effects of reweighting, our framework connects behavioral goals to decisions about how strongly the model learns from each example. Efficient computation brings these decisions into ongoing training. This provides a practical foundation for scalable oversight and control: tracing emerging behavior to the data shaping it and intervening through the learning process itself. Richer differentiable behavioral measurements can extend this framework to new targets for attribution and control.

\clearpage



\begingroup
\urlstyle{same}
\def\UrlBreaks{\do\:\do\-\do\_}
\bibliography{references}

@misc{abdelghafar2026auditinginformationdisclosurellmscale,
  title          = {{Auditing Information Disclosure During LLM-Scale Gradient Descent Using Gradient Uniqueness}},
  author         = {Sleem Abdelghafar and Maryam Aliakbarpour and Chris Jermaine},
  year           = {2026},
  howpublished   = {arXiv preprint arXiv:2510.10902},
  eprint         = {2510.10902},
  archivePrefix  = {arXiv},
  primaryClass   = {cs.LG},
  url            = {https://arxiv.org/abs/2510.10902v2},
}

@misc{kowal2026conceptinfluenceleveraginginterpretability,
  title          = {{Concept Influence: Leveraging Interpretability to Improve Performance and Efficiency in Training Data Attribution}},
  author         = {Matthew Kowal and Goncalo Paulo and Louis Jaburi and Tom Tseng and Lev E McKinney and Stefan Heimersheim and Aaron David Tucker and Adam Gleave and Kellin Pelrine},
  year           = {2026},
  howpublished   = {arXiv preprint arXiv:2602.14869},
  eprint         = {2602.14869},
  archivePrefix  = {arXiv},
  primaryClass   = {cs.AI},
  url            = {https://arxiv.org/abs/2602.14869v1},
}

@inproceedings{jiao2025datelmbenchmarkingdataattribution,
  title          = {{DATE-LM: Benchmarking Data Attribution Evaluation for Large Language Models}},
  author         = {Jiao, Cathy and Pan, Yijun and Xiao, Emily and Sheng, Daisy and Jain, Niket and Zhao, Hanzhang and Dasgupta, Ishita and Ma, Jiaqi and Xiong, Chenyan},
  booktitle      = {Advances in Neural Information Processing Systems},
  volume         = {38},
  publisher      = {Curran Associates, Inc.},
  year           = {2025},
  doi            = {10.52202/085713-5142},
  url            = {https://proceedings.neurips.cc/paper_files/paper/2025/hash/e1ebda145808ca45774993fb67314894-Abstract-Datasets_and_Benchmarks_Track.html},
  note           = {Datasets and Benchmarks Track},
}

@inproceedings{wang2025datashapleytrainingrun,
  title          = {{Data Shapley in One Training Run}},
  author         = {Wang, Jiachen (Tianhao) and Mittal, Prateek and Song, Dawn and Jia, Ruoxi},
  booktitle      = {International Conference on Learning Representations},
  pages          = {12358--12395},
  year           = {2025},
  url            = {https://proceedings.iclr.cc/paper_files/paper/2025/hash/20fdaf67581e6d7157376d1ed584040a-Abstract-Conference.html},
}

@misc{ding2026inrundatashapleyadam,
  title          = {{In-Run Data Shapley for Adam Optimizer}},
  author         = {Meng Ding and Zeqing Zhang and Di Wang and Lijie Hu},
  year           = {2026},
  howpublished   = {arXiv preprint arXiv:2602.00329},
  eprint         = {2602.00329},
  archivePrefix  = {arXiv},
  primaryClass   = {cs.LG},
  url            = {https://arxiv.org/abs/2602.00329v4},
  note           = {Earlier version presented at the 3rd DATA-FM Workshop at ICLR 2026 (non-archival)},
}

@misc{ilyas2025magicnearoptimaldataattribution,
  title          = {{MAGIC: Near-Optimal Data Attribution for Deep Learning}},
  author         = {Andrew Ilyas and Logan Engstrom},
  year           = {2025},
  howpublished   = {arXiv preprint arXiv:2504.16430},
  eprint         = {2504.16430},
  archivePrefix  = {arXiv},
  primaryClass   = {cs.LG},
  url            = {https://arxiv.org/abs/2504.16430v1},
}

@misc{grosse2023studyinglargelanguagemodel,
  title          = {{Studying Large Language Model Generalization with Influence Functions}},
  author         = {Roger Grosse and Juhan Bae and Cem Anil and Nelson Elhage and Alex Tamkin and Amirhossein Tajdini and Benoit Steiner and Dustin Li and Esin Durmus and Ethan Perez and Evan Hubinger and Kamil{\.e} Luko{\v{s}}i{\={u}}t{\.e} and Karina Nguyen and Nicholas Joseph and Sam McCandlish and Jared Kaplan and Samuel R. Bowman},
  year           = {2023},
  howpublished   = {arXiv preprint arXiv:2308.03296},
  eprint         = {2308.03296},
  archivePrefix  = {arXiv},
  primaryClass   = {cs.LG},
  url            = {https://arxiv.org/abs/2308.03296v1},
}

@inproceedings{xia2024lessselectinginfluentialdata,
  title          = {{{LESS}: Selecting Influential Data for Targeted Instruction Tuning}},
  author         = {Xia, Mengzhou and Malladi, Sadhika and Gururangan, Suchin and Arora, Sanjeev and Chen, Danqi},
  booktitle      = {Proceedings of the 41st International Conference on Machine Learning},
  volume         = {235},
  series         = {Proceedings of Machine Learning Research},
  pages          = {54104--54132},
  publisher      = {PMLR},
  year           = {2024},
  url            = {https://proceedings.mlr.press/v235/xia24c.html},
}

@inproceedings{pruthi2020estimatingtrainingdatainfluence,
  title          = {{Estimating Training Data Influence by Tracing Gradient Descent}},
  author         = {Pruthi, Garima and Liu, Frederick and Kale, Satyen and Sundararajan, Mukund},
  booktitle      = {Advances in Neural Information Processing Systems},
  volume         = {33},
  pages          = {19920--19930},
  publisher      = {Curran Associates, Inc.},
  year           = {2020},
  url            = {https://proceedings.neurips.cc/paper/2020/hash/e6385d39ec9394f2f3a354d9d2b88eec-Abstract.html},
}

@inproceedings{koh2020understandingblackboxpredictionsinfluence,
  title          = {{Understanding Black-box Predictions via Influence Functions}},
  author         = {Pang Wei Koh and Percy Liang},
  booktitle      = {Proceedings of the 34th International Conference on Machine Learning},
  volume         = {70},
  series         = {Proceedings of Machine Learning Research},
  pages          = {1885--1894},
  publisher      = {PMLR},
  year           = {2017},
  url            = {https://proceedings.mlr.press/v70/koh17a.html},
}

@misc{chen2025personavectorsmonitoringcontrolling,
  title          = {{Persona Vectors: Monitoring and Controlling Character Traits in Language Models}},
  author         = {Runjin Chen and Andy Arditi and Henry Sleight and Owain Evans and Jack Lindsey},
  year           = {2025},
  howpublished   = {arXiv preprint arXiv:2507.21509},
  eprint         = {2507.21509},
  archivePrefix  = {arXiv},
  primaryClass   = {cs.CL},
  url            = {https://arxiv.org/abs/2507.21509v3},
}

@inproceedings{loshchilov2019decoupledweightdecayregularization,
  title          = {{Decoupled Weight Decay Regularization}},
  author         = {Ilya Loshchilov and Frank Hutter},
  booktitle      = {International Conference on Learning Representations},
  year           = {2019},
  url            = {https://openreview.net/forum?id=Bkg6RiCqY7},
}

@inproceedings{hu2021loralowrankadaptationlarge,
  title          = {{LoRA: Low-Rank Adaptation of Large Language Models}},
  author         = {Edward J. Hu and Yelong Shen and Phillip Wallis and Zeyuan Allen-Zhu and Yuanzhi Li and Shean Wang and Lu Wang and Weizhu Chen},
  booktitle      = {International Conference on Learning Representations},
  year           = {2022},
  url            = {https://openreview.net/forum?id=nZeVKeeFYf9},
}

@inproceedings{dettmers20228bitoptimizersblockwisequantization,
  title          = {{8-bit Optimizers via Block-wise Quantization}},
  author         = {Tim Dettmers and Mike Lewis and Sam Shleifer and Luke Zettlemoyer},
  booktitle      = {International Conference on Learning Representations},
  year           = {2022},
  url            = {https://openreview.net/forum?id=shpkpVXzo3h},
}

@misc{kalajdzievski2023rankstabilizationscalingfactor,
  title          = {{A Rank Stabilization Scaling Factor for Fine-Tuning with LoRA}},
  author         = {Damjan Kalajdzievski},
  year           = {2023},
  howpublished   = {arXiv preprint arXiv:2312.03732},
  eprint         = {2312.03732},
  archivePrefix  = {arXiv},
  primaryClass   = {cs.CL},
  url            = {https://arxiv.org/abs/2312.03732v1},
}

@misc{deng2026faithfultrajectorybaseddataattribution,
  title          = {{How Faithful Is Trajectory-Based Data Attribution? Error Sources, Remedies, and Practical Guidelines}},
  author         = {Junwei Deng and Pingbang Hu and Suliang Jin and Hao Lu and Jiachen T. Wang and Shichang Zhang and Jiaqi W. Ma},
  year           = {2026},
  howpublished   = {arXiv preprint arXiv:2605.18814},
  eprint         = {2605.18814},
  archivePrefix  = {arXiv},
  primaryClass   = {cs.LG},
  url            = {https://arxiv.org/abs/2605.18814v1},
}

@inproceedings{park2023trakattributingmodelbehavior,
  title          = {{{TRAK}: Attributing Model Behavior at Scale}},
  author         = {Park, Sung Min and Georgiev, Kristian and Ilyas, Andrew and Leclerc, Guillaume and Madry, Aleksander},
  booktitle      = {Proceedings of the 40th International Conference on Machine Learning},
  volume         = {202},
  series         = {Proceedings of Machine Learning Research},
  pages          = {27074--27113},
  publisher      = {PMLR},
  year           = {2023},
  url            = {https://proceedings.mlr.press/v202/park23c.html},
}

@inproceedings{tailor2026bayesianinformationtheoreticapproachdata,
  title          = {{A Bayesian Information-Theoretic Approach to Data Attribution}},
  author         = {Tailor, Dharmesh and Felicioni, Nicol\`{o} and Ciosek, Kamil},
  booktitle      = {Proceedings of The 29th International Conference on Artificial Intelligence and Statistics},
  volume         = {300},
  series         = {Proceedings of Machine Learning Research},
  pages          = {1936--1944},
  publisher      = {PMLR},
  year           = {2026},
  url            = {https://proceedings.mlr.press/v300/tailor26a.html},
}

@book{cover2006elements,
  title          = {{Elements of Information Theory}},
  author         = {Cover, Thomas M. and Thomas, Joy A.},
  edition        = {2nd},
  publisher      = {John Wiley \& Sons},
  year           = {2006},
  doi            = {10.1002/047174882X},
  url            = {https://doi.org/10.1002/047174882X},
}

@manual{IMM2012-03274,
  title          = {{The Matrix Cookbook}},
  author         = {K. B. Petersen and M. S. Pedersen},
  organization   = {Technical University of Denmark},
  year           = {2012},
  url            = {https://www2.imm.dtu.dk/pubdb/pubs/3274-full.html},
  note           = {Version 20121115},
}

@inproceedings{NEURIPS2024_ed165f2f,
  title          = {{GREATS: Online Selection of High-Quality Data for LLM Training in Every Iteration}},
  author         = {Wang, Jiachen T. and Wu, Tong and Song, Dawn and Mittal, Prateek and Jia, Ruoxi},
  booktitle      = {Advances in Neural Information Processing Systems},
  volume         = {37},
  pages          = {131197--131223},
  publisher      = {Curran Associates, Inc.},
  year           = {2024},
  doi            = {10.52202/079017-4169},
  url            = {https://proceedings.neurips.cc/paper_files/paper/2024/hash/ed165f2ff227cf36c7e3ef88957dadd9-Abstract-Conference.html},
}

@inproceedings{chang2024scalableinfluencefacttracing,
  title          = {{Scalable Influence and Fact Tracing for Large Language Model Pretraining}},
  author         = {Chang, Tyler and Rajagopal, Dheeraj and Bolukbasi, Tolga and Dixon, Lucas and Tenney, Ian},
  booktitle      = {International Conference on Learning Representations},
  pages          = {40976--40997},
  year           = {2025},
  url            = {https://proceedings.iclr.cc/paper_files/paper/2025/hash/65798a76cc176c29b6bfefe84b0a03ff-Abstract-Conference.html},
}

@article{zhang2024correctinglargelanguagemodel,
  title          = {{Correcting Large Language Model Behavior via Influence Function}},
  author         = {Han Zhang and Zhuo Zhang and Yi Zhang and Yuanzhao Zhai and Hanyang Peng and Yu Lei and Yue Yu and Hui Wang and Bin Liang and Lin Gui and Ruifeng Xu},
  journal        = {Proceedings of the AAAI Conference on Artificial Intelligence},
  volume         = {39},
  number         = {13},
  pages          = {14477--14485},
  year           = {2025},
  doi            = {10.1609/aaai.v39i13.33586},
  url            = {https://ojs.aaai.org/index.php/AAAI/article/view/33586},
}

@inproceedings{hendrycks2021measuringmassivemultitasklanguage,
  title          = {{Measuring Massive Multitask Language Understanding}},
  author         = {Dan Hendrycks and Collin Burns and Steven Basart and Andy Zou and Mantas Mazeika and Dawn Song and Jacob Steinhardt},
  booktitle      = {International Conference on Learning Representations},
  year           = {2021},
  url            = {https://openreview.net/forum?id=d7KBjmI3GmQ},
}
\bibliographystyle{iclr2027_conference}
\endgroup
\newpage
\appendix

\section{Information, Geometry, and Local Prediction}
\label{app:proofs}

This appendix explains why mutual information yields BGU, how the batch-space computation preserves that score, and when local derivatives predict the effects of reweighting. In Sections~\ref{app:channel-covariance}--\ref{app:structural-proofs}, the training state and response vectors are fixed, so we omit the update index $t$. Each optimizer-aware response $q_j$ (Equ.~\ref{eq:signed-effect}) is a column in $\R^m$; the rows of $Q\in\R^{B\times m}$ are $q_j^\top$. Entropy and information use base-two logarithms. Unless stated otherwise, vector norms are Euclidean and matrix norms are induced operator norms.

\subsection{Where the behavioral covariance comes from}
\label{app:channel-covariance}

To find the information carried by example $j$, separate its tagged response from the other batch contributions in Equ.~\ref{eq:local-information-channel}:
\[
R=q_jZ_j+U_{-j},\qquad
U_{-j}=\sum_{k\ne j}q_kZ_k+\epsilon.
\]
The tags $Z_k\sim\mathcal N(0,1)$ and noise $\epsilon\sim\mathcal N(0,\Sigma_\epsilon)$ are independent and zero-mean. Since $\E[Z_kZ_\ell]$ is one for $k=\ell$ and zero otherwise, $\E[U_{-j}]=0$ and its covariance is
\begin{align}
\operatorname{Cov}(U_{-j})
&=\E[U_{-j}U_{-j}^\top]\notag\\
&=\sum_{k,\ell\ne j}q_k\E[Z_kZ_\ell]q_\ell^\top+\E[\epsilon\epsilon^\top]\notag\\
&=\sum_{k\ne j}q_kq_k^\top+\Sigma_\epsilon
=\Sigma_{-j}.
\label{eq:covariance-derivation}
\end{align}
The cross terms with $\epsilon$ vanish by independence and zero means; cross terms between distinct tags vanish for the same reason. Each outer product is the covariance of a unit-variance tagged response. Thus $\Sigma_{-j}$ is the channel's interference covariance, not a sample covariance of training data.

A linear combination of independent Gaussian variables is Gaussian, so $U_{-j}\sim\mathcal N(0,\Sigma_{-j})$. It is also independent of $Z_j$. This identifies both distributions needed for mutual information:
\begin{equation}
R\sim\mathcal N(0,\Sigma_{-j}+q_jq_j^\top),\qquad
R\mid Z_j=z\sim\mathcal N(q_jz,\Sigma_{-j}).
\label{eq:channel-distributions}
\end{equation}
Conditioning on $Z_j=z$ fixes the contribution $q_jz$, but the other batch contributions and noise remain random. The conditional distribution therefore has a shifted mean and covariance $\Sigma_{-j}$. Both covariance matrices are positive definite because the resolution covariance satisfies $\Sigma_\epsilon\succ0$.

\subsection{From Gaussian entropy to information and BGU}
\label{app:information-derivation}

For a continuous random vector $X$ with density $p_X$, its differential entropy in bits is $h_2(X)=-\E[\log_2p_X(X)]$. We use the standard Gaussian entropy formula \citep{cover2006elements}: if $X\sim\mathcal N(\mu,C)$ in $\R^m$ and $C\succ0$, then
\begin{equation}
h_2(X)=\frac12\log_2\!\left((2\pi e)^m\det C\right).
\label{eq:gaussian-entropy}
\end{equation}
Gaussian entropy depends on the covariance, not the mean. Since the conditional covariance in Equ.~\ref{eq:channel-distributions} is the same for every value of $Z_j$, its entropy is unchanged when averaged over the tag. Subtracting conditional entropy from unconditional entropy gives
\begin{align}
I(Z_j;R)
&=h_2(R)-h_2(R\mid Z_j)\notag\\
&=\frac12\log_2\!\left((2\pi e)^m\det(\Sigma_{-j}+q_jq_j^\top)\right)
-\frac12\log_2\!\left((2\pi e)^m\det\Sigma_{-j}\right)\notag\\
&=\frac12\log_2
\frac{\det(\Sigma_{-j}+q_jq_j^\top)}{\det\Sigma_{-j}}.
\label{eq:information-determinants}
\end{align}

To evaluate the ratio, use the \emph{matrix determinant lemma} \citep[Section~1.2]{IMM2012-03274}: for an invertible square matrix $A$ and compatible column vectors $u,v$,
\begin{equation}
\det(A+uv^\top)=\det(A)\bigl(1+v^\top A^{-1}u\bigr).
\label{eq:determinant-lemma}
\end{equation}
Substituting $A=\Sigma_{-j}$ and $u=v=q_j$ now gives
\[
\frac{\det(\Sigma_{-j}+q_jq_j^\top)}{\det\Sigma_{-j}}
=1+q_j^\top\Sigma_{-j}^{-1}q_j.
\]
Putting this ratio into Equ.~\ref{eq:information-determinants} yields
\[
I(Z_j;R)=\frac12\log_2(1+\BGU_j),\qquad
\BGU_j=q_j^\top\Sigma_{-j}^{-1}q_j,
\]
which is Equ.~\ref{eq:information-bgu}. The inverse interference covariance gives BGU its geometry: responses are compared after directions with greater interference have been downweighted. Conditioning on all the other tags would answer a different question. Their contributions could be subtracted, leaving only resolution noise and giving $I(Z_j;R\mid Z_{-j})=\tfrac12\log_2(1+q_j^\top\Sigma_\epsilon^{-1}q_j)$. This would omit the ambiguity due to the other batch examples. \citet{tailor2026bayesianinformationtheoreticapproachdata} instead measure predictive uncertainty after withholding data; our channel concerns behavioral changes through an optimizer update.

\subsection{Why this information measures distinguishability}
\label{app:capacity}

The capacity interpretation follows by reducing the vector observation to its informative direction. Let $\rho=q_j^\top\Sigma_{-j}^{-1}q_j$. Multiplying by the symmetric inverse square root of $\Sigma_{-j}$ makes the interference covariance equal to the identity:
\[
Y=\Sigma_{-j}^{-1/2}R=a_jZ_j+N,\qquad
 a_j=\Sigma_{-j}^{-1/2}q_j,\qquad N\sim\mathcal N(0,I_m).
\]
This whitening transformation is invertible, so it preserves information. The transformed response $a_j$ has squared length $\rho$. For $q_j\ne0$, its unit direction is $e_j=a_j/\sqrt\rho$, and projecting the observation onto that direction gives
\[
T=e_j^\top Y=\sqrt\rho Z_j+N_1,\qquad N_1=e_j^\top N\sim\mathcal N(0,1).
\]
Coordinates orthogonal to $e_j$ contain no response signal. Their Gaussian noise is independent of both $N_1$ and $Z_j$, so they add no information about the tag. Hence $I(Z_j;R)=I(Z_j;T)$: distinguishing the example in the vector channel is equivalent to observing a scalar signal with signal-to-noise ratio $\rho$.

The same reduction applies if we replace $Z_j$ by any input $X$ independent of the interference and satisfying $\E[X^2]\le1$. In that case $T=\sqrt\rho X+N_1$ has variance at most $1+\rho$. A Gaussian maximizes differential entropy at a fixed variance \citep{cover2006elements}, giving
\begin{align*}
I(X;T)&=h_2(T)-h_2(T\mid X)=h_2(T)-h_2(N_1)\\
&\le\tfrac12\log_2\!\left(2\pi e(1+\rho)\right)
-\tfrac12\log_2(2\pi e)
=\tfrac12\log_2(1+\rho).
\end{align*}
For $X\sim\mathcal N(0,1)$, $T$ is Gaussian with variance $1+\rho$, so equality holds. Hence
\[
\sup_{P_X:\E[X^2]\le1}I(X;q_jX+U_{-j})
=\tfrac12\log_2(1+\rho).
\]
When $q_j=0$, the observation is independent of the input and both sides are zero. This unit-power capacity describes distinguishability across independent uses of the fixed local channel.

\subsection{What uniqueness depends on}
\label{app:structural-proofs}

\paragraph{Redundancy within the batch.}
Hold an example's response $q$ and its interference covariance $\Sigma\succ0$ fixed. Adding another batch example with response $v$ changes the covariance to $\Sigma+vv^\top$. We can calculate the effect on BGU using the \emph{Sherman--Morrison identity} \citep[Section~3.2.4]{IMM2012-03274}:
\begin{equation}
(A+uv^\top)^{-1}
=A^{-1}-\frac{A^{-1}uv^\top A^{-1}}{1+v^\top A^{-1}u},
\label{eq:sherman-morrison}
\end{equation}
provided $A$ is invertible and $1+v^\top A^{-1}u\ne0$. Here, take $A=\Sigma$ and $u=v$; the denominator is positive because $\Sigma\succ0$. Multiplying on the left by $q^\top$ and on the right by $q$ gives
\begin{align*}
q^\top(\Sigma+vv^\top)^{-1}q
&=q^\top\Sigma^{-1}q-
\frac{(q^\top\Sigma^{-1}v)(v^\top\Sigma^{-1}q)}{1+v^\top\Sigma^{-1}v}\\
&=q^\top\Sigma^{-1}q-
\frac{(q^\top\Sigma^{-1}v)^2}{1+v^\top\Sigma^{-1}v}.
\end{align*}
The subtracted term is nonnegative because $\Sigma^{-1}$ is symmetric positive definite. Adding a batch example therefore cannot increase BGU when the original responses and resolution are fixed. It decreases BGU precisely when $q^\top\Sigma^{-1}v\ne0$, that is, when the responses overlap after whitening by the original interference covariance.

For $n$ other examples with the same response $q$ under isotropic noise, write $\Sigma=\lambda I_m+nqq^\top$, with $\lambda>0$. Setting $u=v=\sqrt n\,q$ in Equ.~\ref{eq:sherman-morrison} gives
\[
\BGU=q^\top\Sigma^{-1}q
=\frac{\|q\|^2}{\lambda}
-\frac{n\|q\|^4/\lambda^2}{1+n\|q\|^2/\lambda}
=\frac{\|q\|^2}{\lambda+n\|q\|^2}.
\]
For $n=0$ this is the isolated-response score. If instead all other responses $v_k$ are orthogonal to $q$, then
$\Sigma q=\lambda q+\sum_k v_k(v_k^\top q)=\lambda q$.
Multiplying by $\Sigma^{-1}$ gives $\Sigma^{-1}q=q/\lambda$, so BGU remains $\|q\|^2/\lambda$. Orthogonal responses do not obscure this contribution at the chosen isotropic resolution.

\paragraph{Units and coordinates.}
Let $L$ be invertible and transform both responses and resolution by $q'_k=Lq_k$ and $\Sigma'_\epsilon=L\Sigma_\epsilon L^\top$. Expanding the transformed covariance yields
\[
\Sigma'_{-j}=L\Sigma_\epsilon L^\top+\sum_{k\ne j}Lq_kq_k^\top L^\top
=L\Sigma_{-j}L^\top.
\]
The inverse is $L^{-\top}\Sigma_{-j}^{-1}L^{-1}$, as direct multiplication verifies. Therefore
\[
(q'_j)^\top(\Sigma'_{-j})^{-1}q'_j
=q_j^\top L^\top L^{-\top}\Sigma_{-j}^{-1}L^{-1}Lq_j
=\BGU_j.
\]
Information is unchanged because it is a function of BGU. To preserve the signed scores as well, transform the orientation to $a'=L^{-\top}a$; then $(a')^\top q'_j=a^\top q_j$. Keeping isotropic noise after an arbitrary rescaling would measure distinguishability at a different resolution.

\subsection{Deriving the batch-space computation}
\label{app:batch-space-proof}

The goal is to avoid a separate inverse for each example's interference covariance. For isotropic resolution, define $K=QQ^\top\in\R^{B\times B}$ and $M=Q^\top Q+\lambda I_m\in\R^{m\times m}$. The covariance $M$ includes every example, whereas $M_{-j}=M-q_jq_j^\top$ excludes $j$. Positive $\lambda$ makes $M$, $M_{-j}$, and $K+\lambda I_B$ invertible even when $Q$ is rank deficient. We first compute the scores using $M$, then remove each example's self-contribution.

To express the shared covariance calculation in batch space, expand
\begin{align*}
(K+\lambda I_B)Q
&=(QQ^\top)Q+\lambda Q\\
&=Q(Q^\top Q)+\lambda Q
=Q(Q^\top Q+\lambda I_m)=QM.
\end{align*}
Multiplying on the left by $(K+\lambda I_B)^{-1}$ and on the right by $M^{-1}$ yields the \emph{push-through identity}
\[
(K+\lambda I_B)^{-1}Q=QM^{-1}
=Q(Q^\top Q+\lambda I_m)^{-1}.
\]
Multiplying on the right by $Q^\top$ and substituting $QQ^\top=K$ gives
\begin{equation}
H:=QM^{-1}Q^\top=(K+\lambda I_B)^{-1}K
=K(K+\lambda I_B)^{-1}.
\label{eq:kernel-pushthrough}
\end{equation}
The last equality follows because either product equals $I_B-\lambda(K+\lambda I_B)^{-1}$. It also explains why Algorithm~\ref{alg:bgu} solves $(K+\lambda I_B)H=K$. Its diagonal satisfies $h_j=H_{jj}=q_j^\top M^{-1}q_j$.

The diagonal $h_j$ still includes example $j$ in the covariance. To recover the desired score, define $\rho_j=q_j^\top M_{-j}^{-1}q_j=\BGU_j$ and apply Sherman--Morrison to $M=M_{-j}+q_jq_j^\top$:
\[
M^{-1}=M_{-j}^{-1}
-\frac{M_{-j}^{-1}q_jq_j^\top M_{-j}^{-1}}{1+\rho_j}.
\]
Multiplying by $q_j^\top$ and $q_j$ gives
\[
h_j=\rho_j-\frac{\rho_j^2}{1+\rho_j}
=\frac{\rho_j}{1+\rho_j}.
\]
Since $\rho_j\ge0$ is finite, $0\le h_j<1$. Rearranging $h_j(1+\rho_j)=\rho_j$ yields
\[
\BGU_j=\rho_j=\frac{h_j}{1-h_j},
\]
which recovers every leave-one-example-out score from one batch-space solve. Since the derivative of $\tfrac12\log_2(1+\rho)$ is $1/[2\ln2(1+\rho)]>0$, information and BGU rank individual occurrences identically. Sums of these scores need not have identical rankings.

\paragraph{General resolution and coordinate chunks.}
For $\Sigma_\epsilon\succ0$, define $\bar q_j=\Sigma_\epsilon^{-1/2}q_j$ and put these whitened responses in the rows of $\bar Q$. Factoring the covariance gives
\[
\Sigma_{-j}=\Sigma_\epsilon^{1/2}
\left(I_m+\sum_{k\ne j}\bar q_k\bar q_k^\top\right)
\Sigma_\epsilon^{1/2},
\quad
\BGU_j=\bar q_j^\top
\left(I_m+\sum_{k\ne j}\bar q_k\bar q_k^\top\right)^{-1}\bar q_j.
\]
The preceding batch-space derivation therefore applies to $\bar Q$ with ridge one. In the isotropic case we can instead use $Q$ directly with ridge $\lambda$, as in Algorithm~\ref{alg:bgu}.

If the target coordinates are partitioned into disjoint chunks, then $Q=[Q_{C_1}\;Q_{C_2}\;\cdots]$. Block multiplication gives $QQ^\top=\sum_C Q_CQ_C^\top$ and $Qa=\sum_CQ_Ca_C$. These identities justify accumulating both the kernel and the signed projections one chunk at a time. No cross-coordinate products are missing: each Gram entry is the sum of products at matching coordinates.

\paragraph{Signed and accumulated scores.}
We use $\operatorname{sign}(0)=0$. Thus an occurrence with zero projected response has zero signed score even if it has nonzero unsigned information in other directions. For an identity $i$, let $\mathcal O_i$ denote its occurrences. Linearity of projection gives
\[
a^\top Q_i=a^\top\sum_{(t,j)\in\mathcal O_i}q_{t,j}
=\sum_{(t,j)\in\mathcal O_i}r_{t,j},\qquad
C_i=\operatorname{sign}\!\left(\sum_{(t,j)\in\mathcal O_i}r_{t,j}\right)
\sum_{(t,j)\in\mathcal O_i}\mathcal I_{t,j}.
\]
Thus the corpus score in Equ.~\ref{eq:trajectory-aggregation} uses the sign of the summed response and the sum of unsigned information. This differs from summing signed information when an example changes direction across occurrences: opposite signs can cancel in the net response without canceling its accumulated information. Neither sum is joint mutual information across the trajectory.

\subsection{When a local response predicts a finite intervention}
\label{app:finite-proof}

The response is a derivative at unit weights; control applies finite changes. To connect them, fix a log-weight direction $z\in\R^B$ and let $f_t(s;z)=b(F_t(e^{sz}))$. Varying the scalar $s$ follows that direction from ordinary training. The chain rule at $s=0$ gives
\[
f'_t(0;z)=\sum_j
\left.\frac{\partial b(F_t(e^{\mathbf s}))}{\partial s_j}\right|_{\mathbf s=0}z_j
=\sum_jq_{t,j}z_j.
\]
Taking $z=\mathbf1$ proves the uniform-reweighting identity
\begin{equation}
\sum_jq_{t,j}=\left.\frac{d}{ds}b(F_t(e^s\mathbf1))\right|_{s=0}.
\label{eq:linear-completeness}
\end{equation}
This identity describes the derivative when every example's weight changes together. It does not equate the sum with an ordinary finite training update.

The finite prediction error depends on how quickly this derivative changes along the intervention. Suppose $f_t$ is continuously differentiable on $[0,h]$, with $h>0$, and
$\|f'_t(u;z)-f'_t(v;z)\|_2\le M_t(z)|u-v|$ on this interval. The fundamental theorem of calculus gives
\[
f_t(h;z)-f_t(0;z)-h f'_t(0;z)
=\int_0^h\bigl[f'_t(u;z)-f'_t(0;z)\bigr]du.
\]
The triangle inequality bounds the remainder by
$\int_0^h M_t(z)u\,du=M_t(z)h^2/2$.
Substituting the response derivative and dividing by $h$ yields
\begin{equation}
\left\|\frac{f_t(h;z)-f_t(0;z)}h-\sum_jq_{t,j}z_j\right\|_2
\le\frac{M_t(z)h}{2}.
\label{eq:finite-bound}
\end{equation}
For positive weights $\mathbf w$, take $z=\log\mathbf w$ and $h=1$. The linear term is $\sum_jq_{t,j}\log w_j$. The controller uses the affine scale $p(\theta)=\kappa a^\top b(\theta)+c$ (Appendix~\ref{app:feedback}). Applying it gives Equ.~\ref{eq:control-predictor}: the offset $c$ cancels, and the scalar error is at most $\kappa\|a\|_2M_t(z)/2$. This bound requires the stated local smoothness; clipping boundaries and rounded kernels need finite-effect checks.

\paragraph{Error from reusing the target derivative.}
Let $J(\theta)=D_\theta b(\theta)$, $A_t$ be the current optimizer derivative from Section~\ref{sec:algorithm}, and $\tau(t)$ the last target refresh. The exact and reused responses differ only in their target Jacobian:
\[
q_{t,j}=J(\theta_{t+1}^{\rm ord})A_tg_{t,j}/B,
\qquad
\widetilde q_{t,j}=J(\theta_{\tau(t)})A_tg_{t,j}/B.
\]
Assume $J$ is $L_b$-Lipschitz in operator norm between these states. Subtracting the expressions and applying the product-norm inequality gives
\begin{align*}
\|\widetilde q_{t,j}-q_{t,j}\|_2
&=B^{-1}\|[J(\theta_{\tau(t)})-J(\theta_{t+1}^{\rm ord})]A_tg_{t,j}\|_2\\
&\le B^{-1}\|J(\theta_{\tau(t)})-J(\theta_{t+1}^{\rm ord})\|_{\rm op}\|A_tg_{t,j}\|_2.
\end{align*}
Applying the Lipschitz assumption then yields
\begin{equation}
\|\widetilde q_{t,j}-q_{t,j}\|_2
\le\frac{L_b}{B}\|\theta_{t+1}^{\rm ord}-\theta_{\tau(t)}\|_2\|A_tg_{t,j}\|_2.
\label{eq:window-bound}
\end{equation}
Even when targets are refreshed at every step, $\tau(t)=t$, they are evaluated before the update rather than at its ordinary endpoint. The bound includes that difference as well as additional drift between refreshes. For a finite intervention, the extra prediction error is at most $\sum_j|\log w_j|\,\|\widetilde q_{t,j}-q_{t,j}\|_2$. This error adds to the finite-reweighting remainder in Equ.~\ref{eq:finite-bound}.

For comparison, linearizing the ordinary target change is a separate Taylor expansion. Set $\Delta\theta=\theta_{t+1}^{\rm ord}-\theta_t$. If $J$ is $L_b$-Lipschitz along that parameter segment, then
\begin{align*}
b(\theta_{t+1}^{\rm ord})-b(\theta_t)-J(\theta_t)\Delta\theta
&=\int_0^1[J(\theta_t+s\Delta\theta)-J(\theta_t)]\Delta\theta\,ds,\\
\|b(\theta_{t+1}^{\rm ord})-b(\theta_t)-J(\theta_t)\Delta\theta\|_2
&\le\int_0^1 L_bs\|\Delta\theta\|_2^2\,ds
=\tfrac12L_b\|\Delta\theta\|_2^2.
\end{align*}
The three errors have different sources: finite reweighting, a reused target derivative, and linearization across the ordinary parameter update.

\section{The Behavioral BS-Ghost Implementation}
\label{app:algorithm-details}

\subsection{Sharing the optimizer derivative}
\label{app:optimizer-transition}

Let $P$ be the number of trainable parameters. With the starting parameters, batch, and randomness fixed, each example gradient $g_j\in\R^P$ is fixed while its weight varies. We use the weighted mean loss $B^{-1}\sum_j w_j\ell_j$, whose denominator stays $B$ rather than changing with the weights. Here $\ell_j$ includes the training loss's fixed normalization. Its gradient and log-weight derivative are
\[
\bar g(\mathbf s)=\frac1B\sum_j e^{s_j}g_j,
\qquad
\left.\frac{\partial\bar g(\mathbf s)}{\partial s_j}\right|_{\mathbf s=0}
=\frac{e^0g_j}{B}=\frac{g_j}{B}.
\]
For token-normalized training, let $T_j$ be the sum of example $j$'s response-token losses and $N$ the number of response tokens in the complete accumulation group. Defining $\ell_j=BT_j/N$ gives
\[
\frac1B\sum_jw_j\ell_j=\frac1N\sum_jw_jT_j,
\qquad \frac{g_j}{B}=\frac{\nabla_\theta T_j}{N}.
\]
Both $B$ and $N$ remain fixed as weights vary. After mixed-precision loss unscaling, the captured $\nabla_\theta T_j/N$ is exactly the weight-to-gradient factor $g_j/B$ used below.

Let $G_t(g)$ return the parameters after an optimizer step with aggregate gradient $g$ and fixed pre-step state. Then $F_t(e^{\mathbf s})=G_t(\bar g(\mathbf s))$. Define $A_t=DG_t(\bar g(0))\in\R^{P\times P}$ and $J_{t+1}^{\rm ord}=D_\theta b(F_t(\mathbf1))\in\R^{m\times P}$. The chain rule follows the weight change through the aggregate gradient, the optimizer, and the behavioral measurement:
\begin{equation}
q_{t,j}
=\underbrace{J_{t+1}^{\rm ord}}_{\substack{\text{parameters}\\\text{to behavior}}}\;
\underbrace{A_t}_{\substack{\text{gradient}\\\text{to parameters}}}\;
\underbrace{\frac{g_j}{B}}_{\substack{\text{log weight}\\\text{to gradient}}}.
\label{eq:optimizer-chain-rule}
\end{equation}
For target coordinate $r$, let $u_r=\nabla_\theta b_r(\theta_{t+1}^{\rm ord})\in\R^P$. Transposing the first two factors gives
$q_{t,j,r}=u_r^\top A_tg_j/B=(A_t^\top u_r/B)^\top g_j$.
The transformed target $v_{t,r}=A_t^\top u_r/B$ can therefore be shared by every example. BS-Ghost computes this matrix--vector action without constructing the dense $P\times P$ optimizer Jacobian.

For minibatch SGD with gradients $g_{t,j}=\nabla_\theta\ell(\theta_t,d_{t,j})$ and learning rate $\eta_t$, the update is $F_t(\mathbf w)=\theta_t-\eta_t B^{-1}\sum_jw_jg_{t,j}$. Differentiating $e^{s_j}$ at zero gives $-\eta_tg_{t,j}/B$; applying the behavioral Jacobian yields
\begin{equation}
q_{t,j}=-\frac{\eta_t}{B}D_\theta b(\theta_{t+1}^{\rm ord})g_{t,j}.
\label{eq:sgd-process-effect}
\end{equation}
The target Jacobian is evaluated at the ordinary next state, the composed map's output at $\mathbf s=0$.

\paragraph{AdamW without clipping.}
For AdamW \citep{loshchilov2019decoupledweightdecayregularization}, denote the fixed pre-step first and second moments by $\mu$ and $\nu$. Let $\beta_1,\beta_2$ be their decay factors, $\gamma$ the weight-decay coefficient, and $\epsilon>0$ the denominator stabilizer. Products, squares, division, and square roots below act coordinatewise. With $t+1$ the update count, the transition is
\begin{align*}
\mu^+&=\beta_1\mu+(1-\beta_1)g,&
\nu^+&=\beta_2\nu+(1-\beta_2)g^2,\\
\hat\mu&=\mu^+/(1-\beta_1^{t+1}),&
\hat\nu&=\nu^+/(1-\beta_2^{t+1}),\\
G_t(g)&=(1-\eta_t\gamma)\theta_t
-\eta_t\hat\mu/(\sqrt{\hat\nu}+\epsilon).
\end{align*}
Each output parameter depends only on the matching coordinate of $g$, so $D_t=DG_t(g)$ is diagonal. For coordinate $k$, differentiate the moments first:
\[
\frac{\partial\hat\mu_k}{\partial g_k}
=\frac{1-\beta_1}{1-\beta_1^{t+1}}=:d_\mu,
\qquad
\frac{\partial\hat\nu_k}{\partial g_k}
=\frac{2(1-\beta_2)g_k}{1-\beta_2^{t+1}}=:(d_\nu)_k.
\]
Set $s_k=\sqrt{\hat\nu_k}+\epsilon$. For $\hat\nu_k>0$, use $\partial s_k/\partial g_k=(d_\nu)_k/(2\sqrt{\hat\nu_k})$ in the quotient rule:
\begin{align*}
\frac{\partial}{\partial g_k}\frac{\hat\mu_k}{s_k}
&=\frac{d_\mu s_k-\hat\mu_k\,\partial s_k/\partial g_k}{s_k^2}\\
&=\frac{d_\mu}{\sqrt{\hat\nu_k}+\epsilon}
-\frac{\hat\mu_k(d_\nu)_k}
{2\sqrt{\hat\nu_k}(\sqrt{\hat\nu_k}+\epsilon)^2}.
\end{align*}
Multiplying by $-\eta_t$ yields
\begin{equation}
(D_t)_{kk}=-\eta_t\left[
\frac{d_\mu}{\sqrt{\hat\nu_k}+\epsilon}
-\frac{\hat\mu_k(d_\nu)_k}
{2\sqrt{\hat\nu_k}(\sqrt{\hat\nu_k}+\epsilon)^2}
\right].
\label{eq:adamw-jacobian}
\end{equation}
The first term differentiates the updated first moment; the second accounts for the gradient's effect on the adaptive denominator through the second moment. Freezing that denominator would omit the second term. The weight-decay term above has zero gradient derivative because $\theta_t$ is fixed. If decay is applied after the adaptive step, it instead multiplies the entire transition and its derivative:
\[
G_t^{\rm after}(g)=(1-\eta_t\gamma)
\left[\theta_t-\eta_t\hat\mu/(\sqrt{\hat\nu}+\epsilon)\right],
\qquad DG_t^{\rm after}(g)=(1-\eta_t\gamma)D_t.
\]
Our AdamW8bit path uses this decay order and fixed, dequantized moments. Equ.~\ref{eq:adamw-jacobian} applies to the real-arithmetic transition with positive updated variance.

\paragraph{Adding norm clipping.}
Let $C>0$ be the clipping limit and $\epsilon_c\ge0$ its norm stabilizer. The clipping map is
$c(g)=\min\{1,C/(\|g\|_2+\epsilon_c)\}g$.
On the inactive branch, $Dc(g)=I_P$. On the active branch, write $n_g=\|g\|_2>0$ and $\alpha=C/(n_g+\epsilon_c)$. Using $\nabla_g\|g\|_2=g/n_g$, we obtain $\nabla_g\alpha=-\alpha g/[n_g(n_g+\epsilon_c)]$. Applying the product rule to $c(g)=\alpha g$ gives
\begin{equation}
Dc(g)=\alpha I_P+g(\nabla_g\alpha)^\top
=\alpha\left[I_P-\frac{gg^\top}{n_g(n_g+\epsilon_c)}\right].
\label{eq:clipping-jacobian}
\end{equation}
The optimizer receives $c(\bar g)$, so its full derivative with respect to the original aggregate is $A_t=D_tDc(\bar g)$, with $D_t$ evaluated at $c(\bar g)$. Set $d_r=D_t^\top u_r$ and now let $r_g=\|\bar g\|_2$. Substituting Equ.~\ref{eq:clipping-jacobian} into the shared response gives
\begin{align*}
q_{t,j,r}&=B^{-1}d_r^\top Dc(\bar g)g_j\\
&=\frac{\alpha}{B}\left[
 g_j^\top d_r-
 \frac{(g_j^\top\bar g)(\bar g^\top d_r)}{r_g(r_g+\epsilon_c)}
\right].
\end{align*}
At $\epsilon_c=0$, the correction denominator becomes $\|\bar g\|_2^2$. The contraction requires inner products and one norm over all trainable blocks, including blocks whose target derivative vanishes. In particular, $g_j^\top\bar g$ uses the same ghost identity as Equ.~\ref{eq:ghost-response}, substituting the already materialized aggregate block $\bar g^{(\ell)}$ for $V_{t,r}^{(\ell)}$. No individual parameter gradient is needed.

The Qwen implementation uses $\alpha=C/(n_g+10^{-6})$ when $n_g>C$ and $\widetilde D c=\alpha(I_P-gg^\top/n_g^2)$. It retains the stabilizer only in $\alpha$, so this differs from the exact stabilized derivative in Equ.~\ref{eq:clipping-jacobian}. The finite-update checks assess its numerical agreement.

\subsection{Deriving the ghost contractions}
\label{app:ghost-derivation}

We extend the ghost kernels of \citet{abdelghafar2026auditinginformationdisclosurellmscale} to compute optimizer-aware behavioral responses without materializing full per-example gradients. For an affine trainable block $W^{(\ell)}\in\R^{d_{\rm out}\times d_{\rm in}}$, let $y_{j,s}^{(\ell)}=W^{(\ell)}x_{j,s}^{(\ell)}$ and $\delta_{j,s}^{(\ell)}=\partial\ell_j/\partial y_{j,s}^{(\ell)}$. The chain rule gives the example gradient
\[
g_j^{(\ell)}=\frac{\partial\ell_j}{\partial W^{(\ell)}}
=\sum_s\delta_{j,s}^{(\ell)}(x_{j,s}^{(\ell)})^\top.
\]
Let $V_r^{(\ell)}$ be the matching block of the shared target vector $v_{t,r}$. Decompose $v_{t,r}^\top g_j$ into Frobenius products over parameter blocks. The identity $\langle A,B\rangle_F=\operatorname{tr}(A^\top B)$ gives
\begin{align*}
\langle V_r^{(\ell)},g_j^{(\ell)}\rangle_F
&=\sum_s\operatorname{tr}\!\left((V_r^{(\ell)})^\top
\delta_{j,s}^{(\ell)}(x_{j,s}^{(\ell)})^\top\right)\\
&=\sum_s(x_{j,s}^{(\ell)})^\top(V_r^{(\ell)})^\top\delta_{j,s}^{(\ell)}\\
&=\sum_s(\delta_{j,s}^{(\ell)})^\top V_r^{(\ell)}x_{j,s}^{(\ell)}.
\end{align*}
The second line uses the cyclic trace identity; the last transposes a scalar. Summing over blocks gives Equ.~\ref{eq:ghost-response}. Computing $V_rx$ or $\delta^\top V_r$ first avoids forming the example's full gradient.

When the reference gradient is also factored, let $\bar x$ and $\bar\delta$ denote its activations and errors. Expanding both gradients and applying the same trace identity gives
\begin{align*}
\left\langle\sum_s\delta_{j,s}x_{j,s}^\top,
\sum_v\bar\delta_{r,v}\bar x_{r,v}^\top\right\rangle_F
&=\sum_{s,v}\operatorname{tr}\!\left(x_{j,s}\delta_{j,s}^\top
\bar\delta_{r,v}\bar x_{r,v}^\top\right)\\
&=\sum_{s,v}(\delta_{j,s}^\top\bar\delta_{r,v})(x_{j,s}^\top\bar x_{r,v}).
\end{align*}
Both gradients sum over positions, so their inner product includes every pair $(s,v)$; restricting to $s=v$ would omit terms. A bias uses a constant input factor of one. LoRA's two trainable factors \citep{hu2021loralowrankadaptationlarge} are treated as separate affine parameter blocks. For adaptive optimizers, the reference is the transformed target $v_{t,r}$; a raw target-gradient inner product omits the optimizer derivative. Ghost inner products also appear in GREATS and in-run valuation \citep{NEURIPS2024_ed165f2f,wang2025datashapleytrainingrun,ding2026inrundatashapleyadam}.

\subsection{Packing, chunking, and memory}
\label{app:packing-memory}

Packing reduces repeated work without changing the contraction. For LoRA, we can first sum token outer products into small per-example adapter gradients and reuse them across behavioral targets. Qwen groups compatible blocks and retains these gradients and cached targets on the GPU until the effective batch is scored. Target prompts are batched and evaluated only through the measured layer. DATE-LM instead contracts and releases one LoRA block at a time to preserve its reference arithmetic.

The response buffers require $Bc$ values for a chunk of $c$ coordinates and $B^2$ for the running kernel. This $O(Bc+B^2)$ storage is only part of the memory cost. Final contractions depend on the complete aggregate gradient through clipping and the optimizer derivative. A factor-based implementation must retain or recompute the activations and backward errors until that gradient is available. Retaining them costs $O(\sum_\ell BL_\ell(d_{{\rm in},\ell}+d_{{\rm out},\ell}))$ values, where $L_\ell$ is the number of token positions. The optimized LoRA path instead retains $O(BP_{\rm adapter})$ per-example adapter-gradient values, plus cached targets and temporary grouping buffers. It never forms gradients of the frozen base weights. The large-model experiments use LoRA; Appendix~\ref{app:numerical-validation} also checks models without LoRA. Offloading or recomputation may be needed for full-parameter LLM training and has not been benchmarked here.

Each of the $m$ coordinates contributes an outer product of length $B$, giving $O(B^2m)$ kernel accumulation; the dense $B\times B$ solve costs $O(B^3)$. Contraction cost also depends on layer dimensions and sequence lengths. Over $T$ consecutively attributed updates, refreshing targets every $W$ updates reduces target evaluations from $T$ to approximately $T/W$ while scoring every occurrence. Bulk transfers to preallocated host buffers avoid per-score synchronization; timing includes transfers and final synchronization.

\paragraph{Precision of the batch-space solve.}
Qwen uses FP32 for the optimized contractions and FP64 for response accumulation, the batch kernel, the solve, and extraction of $h_j/(1-h_j)$. The model's lower precision does not carry over to this solve. Although positive resolution ensures $h_j<1$ analytically, finite precision could round it to one and make the denominator vanish. The measured path uses no clamp. For nonzero $K$ and $\lambda=\alpha\operatorname{tr}(K)/B$, positive semidefiniteness gives $1-h_j\ge\lambda/(\lambda+\|K\|_{\rm op})\ge\alpha/(B+\alpha)$, since $h_j\le\|H\|_{\rm op}$ and $\|K\|_{\rm op}\le\operatorname{tr}(K)$. At the experiment's $\alpha=10^{-4}$ and $B=16$, this separation from one is readily resolved in FP64. All-zero response batches receive zero scores.

\section{Numerical Fidelity and Target Reuse}
\label{app:numerical-validation}

Efficient computation is useful only if it preserves the quantities on which attribution and feedback depend. We therefore check the contractions and batch-space identities, the prediction of finite reweighting, and the effect of reusing target derivatives.

\paragraph{Algebraic fidelity.}
We compare BS-Ghost with directly materialized gradients on linear, MLP, and transformer models. The checks cover affine and LoRA blocks, biases, cross-position terms, coordinate chunking, and several batch and target sizes. In double precision, maximum absolute errors were below $10^{-12}$ for responses, BGU, and information. In single precision, the largest relative contraction error was below $0.004\%$. A batch containing one example also recovers response energy divided by resolution, as the theory requires.

\paragraph{Finite reweighting.}
On MLP and transformer trajectories, explicit SGD and AdamW derivatives agree with automatic differentiation to numerical precision. We then compare their predictions with actual log-weight perturbations of size $10^{-5}$. Raw-gradient prediction has median relative error of about $0.13\%$ for SGD and $56\%$ for AdamW; including the optimizer reduces the error to below $0.001\%$. Thus differentiating the actual optimizer matters to prediction before any adjustment for batch interference. Appendix~\ref{app:incident-control} checks finite interventions with Qwen.

\paragraph{Target freshness.}
\label{app:window-results}
Reusing target derivatives saves backward passes at the cost of the approximation bounded in Equ.~\ref{eq:window-bound}. Example factors and the optimizer derivative remain current at every attributed update. On 816 Qwen development examples, refreshing targets every four updates gives a BGU rank correlation of $0.97$ with fresh targets. The corresponding 10\%, 25\%, and 50\% removal sets overlap by about 83\%, 93\%, and 98\%. These development checks determined the refresh window before the removal experiment.

\section{Experimental Targets, Data, and Evaluation}
\label{app:protocol-fidelity}\label{app:behavior}\label{app:repro-table}

We give the protocols for data removal, matched score computation, and attribution under batch interference. Appendix~\ref{app:temporal-analysis} covers temporal attribution and feedback.

\subsection{Qwen information-based removal}
\label{app:concept-influence}

To test whether examples identified during training shape final behavior, we train Qwen2.5-7B-Instruct on each 9,000-example corpus, remove examples according to their scores, and retrain. Training uses rank-32 RS-LoRA \citep{kalajdzievski2023rankstabilizationscalingfactor} and AdamW8bit \citep{dettmers20228bitoptimizersblockwisequantization}, with learning rate $10^{-5}$, five warmup steps, sequence length 2,048, and effective batch size 16. One pass takes 563 updates.

For attribution, we follow Concept Influence and Persona Vectors \citep{kowal2026conceptinfluenceleveraginginterpretability,chen2025personavectorsmonitoringcontrolling} and measure behavior along a fixed persona direction $v$. Each coordinate $b_r(\theta)=v^\top h_{20}(x_r;\theta)$ projects the layer-20 activation at the final token of a development prompt. These 15 prompts define the internal measurement used to score training examples; twenty separate prompts evaluate generated answers.

The information calculation uses isotropic resolution $\lambda_t=10^{-4}\operatorname{tr}(K_t)/B$, with underflow protection. This value is fixed within each local channel; an all-zero batch receives zero information. Because each training example occurs once, its corpus score in Equ.~\ref{eq:trajectory-aggregation} is simply its signed information, $C_i=S_{t,j}$.

We evaluate removal through generated answers rather than the internal projection used for attribution. For each removal fraction, three matched training runs each produce ten answers to all twenty evaluation prompts: 200 generations per run and 600 per condition. The judge \mbox{\texttt{gpt-4.1-mini-2025-04-14}} scores evil and coherence using the benchmark's probability-weighted mean of valid integer ratings from 0 to 100. Student 95\% intervals describe variation across training runs; score intervals are truncated at zero.

\begin{table}[!t]
\caption{\textbf{Behavioral change and utility after information-selected removal.} Each run contributes 200 generations. Lower evil and higher coherence are better. $\Delta$NLL is filtered minus matched unfiltered negative log-likelihood; lower is better. These runs give Figure~\ref{fig:phase4a-remove-most}'s means and uncertainty.}
\label{tab:phase4a-seeds}
\centering
\scriptsize
\setlength{\tabcolsep}{3.2pt}
\begin{tabular}{@{}llrrrrr@{}}
\toprule
Dataset & Removed & Evil run 1 & Evil run 2 & Evil run 3 & Coherence mean & $\Delta$NLL \\
\midrule
\multirow{7}{*}{Opinions}
 & 0\% & 27.15 & 20.37 & 21.97 & 88.24 & 0 \\
 & 10\% & 17.45 & 18.61 & 22.46 & 89.35 & .0089 \\
 & 25\% & 9.40 & 12.97 & 11.34 & 92.60 & .0276 \\
 & 50\% & 1.96 & 4.45 & 3.99 & 95.93 & .0711 \\
 & 75\% & .85 & 1.23 & 2.37 & 96.06 & .1302 \\
 & 90\% & 1.47 & 2.13 & 1.79 & 96.00 & .2362 \\
 & 100\% & 0 & 0 & 0 & 99.12 & 1.1897 \\
\midrule
\multirow{7}{*}{Medical}
 & 0\% & 22.01 & 25.73 & 18.34 & 90.24 & 0 \\
 & 10\% & 16.01 & 20.57 & 19.05 & 90.92 & .0099 \\
 & 25\% & 10.77 & 12.81 & 9.32 & 94.25 & .0292 \\
 & 50\% & 6.43 & 6.62 & 6.98 & 95.56 & .0851 \\
 & 75\% & 1.59 & 2.43 & 3.35 & 96.53 & .1412 \\
 & 90\% & .89 & .08 & 1.15 & 96.70 & .2263 \\
 & 100\% & 0 & 0 & 0 & 99.13 & .9202 \\
\midrule
\multirow{7}{*}{GSM8K}
 & 0\% & 9.98 & 10.41 & 15.28 & 92.20 & 0 \\
 & 10\% & 2.90 & 2.75 & 1.48 & 95.44 & .0035 \\
 & 25\% & 0 & .30 & 0 & 97.46 & .0118 \\
 & 50\% & 0 & $<.01$ & 0 & 97.50 & .0237 \\
 & 75\% & 0 & .15 & 0 & 97.49 & .0399 \\
 & 90\% & 0 & 0 & 0 & 97.39 & .0809 \\
 & 100\% & 0 & 0 & 0 & 99.10 & .7669 \\
\bottomrule
\end{tabular}
\end{table}

Across paired runs, 50\% removal reduces evil by $19.69$ points $[7.60,31.79]$ on Opinions and $15.35$ $[5.72,24.98]$ on Medical; on GSM8K, 10\% removal gives a reduction of $9.51$ $[0.26,18.76]$. Coherence generally improves, although extensive removal raises held-out negative log-likelihood (NLL). Table~\ref{tab:phase4a-seeds} reports both outcomes; NLL averages response-token loss over 500 benign and 500 misaligned held-out examples per corpus. Other methods' curves in Figure~\ref{fig:phase4a-remove-most} are digitized external references, with unmatched software, hardware, and evaluation conditions.

\subsection{DATE-LM: matched attribution computation}
\label{app:datelm}

We compare the cost of computing the same attribution score in DATE-LM. Each pool combines 10,000 UltraChat examples with 97 ToxicChat, 66 XSTest-response, or 70 JailbreakBench examples. We train Pythia-1B with rank-8 query/key/value LoRA for five epochs, using sequence length 1,024, batch size four, and AdamW at $10^{-5}$ with a linear schedule and 3\% warmup. Both implementations in Table~\ref{tab:datelm-main} score the same checkpoint and reference answers.

\paragraph{Preserving the score while changing its computation.}
Let $g_i$ be a training-example loss gradient and $h_r$ a reference-answer loss gradient, both over the trainable LoRA parameters. Grad-Dot averages $g_i^\top h_r$ over references; Grad-Sim normalizes the gradients before taking their inner products. DATE-LM's single-checkpoint LESS first transforms candidate gradients using saved Adam moments, projects them and the reference gradients with a fixed 8,192-dimensional Rademacher projection, and averages their cosines \citep{pruthi2020estimatingtrainingdatainfluence,xia2024lessselectinginfluentialdata}. BS-Ghost preserves these definitions, including LESS's nonlinear transformation, while reusing reference calculations and contracting compact gradient factors one LoRA block at a time.

\paragraph{Published and recomputed results.}
DATE-LM's published AUPRC and A6000 times come from its Tables 11 and 14 \citep{jiao2025datelmbenchmarkingdataattribution}; Table~\ref{tab:datelm-main} separates them from our matched V100 results. The public code, checkpoints, and data yield scores that differ from the released tensors, which reproduce the published AUPRC. In particular, ToxicChat's Grad-Dot/Grad-Sim tensors use ten references, whereas the public entrypoint supplies thirty. For the matched V100 comparison, both implementations use the same local configuration and agree within $0.004\%$ relative score error.

\paragraph{Runtime comparison.}
We time the complete post-training scoring pass, including reference preparation, transfers, and final synchronization. BS-Ghost computes Grad-Dot and Grad-Sim in shared passes, from which we derive conservative costs for each score; LESS is timed directly. The implementation also gains speed from caching and length-grouped batches of four, while the reference processes examples individually. This increases peak memory allocation from about 7.0 to 14.0\,GiB for Grad-Dot/Grad-Sim and from 7.4 to 22.5\,GiB for LESS.

\subsection{Attribution Under Batch Interference}
\label{app:distinguishability}

\paragraph{Interventions and recovery.}
Figure~\ref{fig:intro-distinguishability} tests whether attribution scores predict which examples remain identifiable when their effects combine in a batch update.

We train GPT-2 (124M) on constructed entity-to-color facts and evaluate 12 contexts spanning four trained states. Each context fixes the model, optimizer, and a 12-example batch. Every trial restores the same state and independently increases or decreases each example's weight: $z_j\in\{-1,+1\}$ and $w_j=\exp(0.05z_j)$. The weighted loss retains the fixed denominator 12. After executing the update, we measure the change relative to ordinary training, $\Delta b=b(F_t(\mathbf w))-b(F_t(\mathbf1))$, across 48 reference-question cross-entropies. A decoder then tries to recover the weight assignments $z_j$ from these measured loss changes. It receives no attribution scores. Each context supplies 256 training trials, 64 validation trials, and 256 independent test trials.

\paragraph{Decoder training.}
The nonlinear decoder maps the 48 loss changes to twelve weight assignments through a $48\!\to\!32\!\to\!12$ multilayer perceptron, with a $\tanh$ hidden layer and linear output. Each input coordinate is centered by its training mean; all coordinates are then divided by a single training root-mean-square scale. Target assignments are also centered, with their training means restored after prediction. Test trials do not enter either normalization.

We minimize recovery MSE plus $\alpha\operatorname{mean}(\text{weights}^2)$, leaving biases unpenalized. Validation error selects $\alpha=1$ from $\{10^{-4},10^{-2},1\}$ before test evaluation, and we report three initializations. Training uses 400 full-batch epochs of Adam on CPU in FP64, with learning rate $0.01$, $\beta=(0.9,0.999)$, $\epsilon=10^{-8}$, and standard bias correction; weight decay, scheduling, and clipping are disabled. To check dependence on decoder class, we also fit a linear decoder with the same preprocessing and ridge $0.01$: $(X^\top X/n+0.01I)C=X^\top Z/n$, where $X$ contains the normalized loss changes, $Z$ the centered assignments, $n$ the number of training trials, and $C$ the fitted coefficients.

\paragraph{Attribution scores.}
Lower test MSE means easier recovery of an example's weight change. We correlate attribution scores with negative recovery MSE, so a higher Spearman correlation means better ranking of recoverable contributions. Figure~\ref{fig:intro-distinguishability} averages these correlations equally across the twelve contexts and three decoder initializations.

BGU and information induce the same ranking, while our response score is the magnitude $\|q_{t,j}\|_2$. For other methods with one attribution per reference question, we first perform the method's temporal aggregation and then take the norm of the resulting 48 signed coordinates. GREATS supplies a selection order instead, so earlier selections receive a higher rank. All methods use the same reference questions. Except for LinFAC's surrogate modules described below, they operate on the 442,368 parameters in 48 trainable attention-LoRA blocks.

\textbf{MAGIC} differentiates each final reference loss through updates 96--128, propagating derivatives through model parameters and Adam moments \citep{ilyas2025magicnearoptimaldataattribution}. Each candidate occurs four times in this interval, and its score sums the derivatives from all four visits.

Wang et al.'s \textbf{In-Run Data Shapley} approximates the change in reference loss at each update and sums over the same four candidate visits \citep{wang2025datashapleytrainingrun}. The first-order term is $-\eta d_j^\top h$; adding curvature gives $-\eta d_j^\top h+\frac{\eta^2}{2}d_j^\top H\sum_i d_i$. Here $d_j=\nabla\ell_j/12$, $\eta$ is the learning rate, and $h,H$ are the pre-update reference-loss gradient and Hessian. The sum includes example $j$, so the curvature term accounts for its interaction with the whole batch. We evaluate this term through Hessian--vector products, using the authors' SGD-derived utility along Adam training as described in their Appendix~E.2, Remark~7.

\textbf{Adam Shapley}'s \href{https://arxiv.org/html/2602.00329v1}{singleton-update} and \href{https://arxiv.org/html/2602.00329v4}{batch-Jacobian} formulations accumulate over the same candidate visits \citep{ding2026inrundatashapleyadam}. The batch-Jacobian calculation linearizes the paper's Adam map at the observed batch gradient and contracts the resulting example direction with the pre-update reference gradient. Each atomic loss is $\ell_j/12$; the denominator remains fixed when a coalition contains fewer examples.

\textbf{LESS} compares each candidate's Adam-transformed gradient with the reference-loss gradient, following Definition~3.1 of \citet{xia2024lessselectinginfluentialdata}. We project the bias-corrected Adam features and raw query gradients to 8,192 dimensions, compute their cosines, and sum them with learning-rate weights over epoch-end checkpoints 104, 112, 120, and 128.

\textbf{TRAK} uses the correct-answer vocabulary log odds as its model output, together with the loss factor $1-p$, where $p$ is the correct-answer probability \citep{park2023trakattributingmodelbehavior}. At checkpoint 96, we project gradients to 8,192 dimensions and use all 96 candidate copies for whitening. A stable dual solve evaluates the regression with ridge $0.001$.

\textbf{EK-FAC} estimates influence using eigenvalue-corrected Kronecker curvature \citep{grosse2023studyinglargelanguagemodel}. We compute model-Fisher curvature on the shared LoRA blocks and apply damping $10^{-8}$ when inverting it.

\textbf{TrackStar} corrects loss gradients using saved Adam second moments, applies two-sided Gaussian projection, whitens with mixed training/query curvature, and normalizes globally \citep{chang2024scalableinfluencefacttracing}. Eight layer groups contribute 4,096 dimensions each, for 32,768 in total. Training/query mixture weights are $0.10/0.90$, with ridge $\max(10^{-3},10^{-8}\operatorname{tr}R)$ for each mixed curvature $R$. Candidates, references, and checkpoint match TRAK's. The calculation uses our training setup's Adam moments and attention-LoRA parameters in place of the paper's Adafactor moments and all-nonembedding parameter scope.

\textbf{GREATS} selects examples greedily, adjusting their validation-loss utility for interactions with examples already selected \citep{NEURIPS2024_ed165f2f}. At the fixed checkpoint, each candidate starts with its gradient alignment to the mean loss on the 48 references. After a selection, the identity-Hessian correction subtracts that candidate's gradient interactions from the remaining scores, using learning rate $3\times10^{-4}$. Continuing through all twelve candidates gives the complete order used for recovery ranking; exact ties follow candidate order.

\textbf{LinFAC}, LANCET's attribution component, computes sequence-level influence through linear module surrogates \citep{zhang2024correctinglargelanguagemodel}. We use twelve $768\times768$ attention surrogates. Its sequence gradient sums the outer products of output errors and inputs across token positions. Model-Fisher curvature is estimated from 96 candidate copies with one model-sampled answer token each, using mean input factors and summed output-error factors over nonpadding positions. We invert the Kronecker curvature with damping $10^{-8}$ and sum signed query influences across modules; curvature and contractions use FP64. The resulting 7.08 million surrogate coordinates cover attention modules, excluding MLP surrogates. This experiment evaluates LinFAC's attribution scores, without LANCET's retrieval filters or corrective post-training.

\paragraph{Implementations.}
TRAK uses the released projection and normalization routines, and EK-FAC uses Kronfluence. The GREATS selection orders agree with the released greedy function. MAGIC, Shapley, TrackStar, and LinFAC use independent implementations checked against the methods' equations and numerical references. In contexts that replace the intervention batch, trajectory-dependent scores reuse the grouped training trajectory for the same seed. The decoder still measures recovery at the fixed intervention state.

\paragraph{Interpretation.}
An example can move behavior substantially yet be hard to identify when other examples have similar effects. BGU better predicts which contributions remain identifiable: its mean recovery correlation is $0.78$, compared with $0.59$ for our response and $0.58$ for EK-FAC, the strongest external score. BGU also ranks highest with the linear decoder. This supports BGU's handling of attribution ambiguity when examples act together.

The BGU--response difference is $0.20$, with a conditional 95\% interval of $[0.15,0.24]$. We obtain intervals from 2,000 paired bootstrap resamples of whole test trials, holding the trained states and decoders fixed and applying Bonferroni correction across 42 score comparisons. The intervals therefore describe uncertainty within these contexts; their extreme adjusted endpoints have limited Monte Carlo precision. First- and second-order Shapley give similar correlations ($0.498/0.495$), consistent with the small second-order improvement reported by \citet[Appendix~E.2.2]{wang2025datashapleytrainingrun} in a different GPT-2 loss-change approximation experiment.

Panel A's cosines use all 48 response coordinates. Its three arrows are normalized and shown from a fixed viewpoint in the three-dimensional subspace they span. This illustrates directional overlap; the original magnitudes differ, so the panel does not isolate overlap from magnitude.

\section{Temporal Explanation and Feedback}
\label{app:temporal-analysis}

\subsection{Reading signed information over time}

To examine how data order affects behavioral development, we arrange the same 4,500 benign and 4,500 misaligned examples in four orders. Figure~\ref{fig:controlled-order} compares the resulting behavioral state with the contributions made during training. We summarize signed information over a centered 15-update window: for source $c$, $\mathcal W_c(t)$ contains its occurrences from updates $t-7$ through $t+7$. The red curve averages strengthening contributions from misaligned examples, and the blue curve averages weakening contributions from benign examples:
\[
\bar S_{\rm red}(t)=\frac{\sum_{(u,j)\in\mathcal W_{\rm mis}(t)}\max(S_{u,j},0)}{|\mathcal W_{\rm mis}(t)|},
\qquad
\bar S_{\rm blue}(t)=\frac{\sum_{(u,j)\in\mathcal W_{\rm ben}(t)}\min(S_{u,j},0)}{|\mathcal W_{\rm ben}(t)|}.
\]
An occurrence with the opposite sign contributes zero to the numerator but still counts in the denominator. A curve can therefore increase in magnitude because the indicated contributions become more frequent, stronger, or both. Windows with no examples from that source appear as gaps. The curves measure contributions in bits; the accompanying internal readout measures the model's behavioral state.

\subsection{Prediction of information-selected interventions}
\label{app:incident-control}

Before using predictions for repeated feedback, we test whether they describe the effect of a finite weight change. In three Opinions training runs, we intervene every four updates from step 8. Signed information selects up to four examples that strengthen the target behavior. Their weights fall from 1 to $e^{-0.5}$, while the remaining weights stay at one.

We execute ordinary and reweighted updates from the same state, so the difference in their behavioral measurements isolates the intervention. The prediction $\sum_jq_{t,j}\log w_j$ describes that difference across all 15 coordinates. Across 339 interventions, 113 per run, its mean and median cosine with the measured change are $0.84$ and $0.94$. Figure~\ref{fig:intervention-prediction} shows the projection onto $a=\mathbf1/\sqrt{15}$ in raw projection units: RMSE is $0.042$, compared with $0.230$ for predicting no change. These checks of prediction direction and magnitude support using the local predictor repeatedly in the feedback experiment below.

\subsection{Repeated feedback through example weights}
\label{app:feedback}

The feedback comparison follows the Early data order with one matched trajectory per arm. We specify the behavioral target, weight allocation, and evaluation below.

\paragraph{Behavioral scale and activation.}
The internal readout is $x_t=a^\top b(\theta_t)$, with $a=\mathbf1/\sqrt{15}$. We express control thresholds on a normalized scale, $p_t=100(x_t-z_{\rm safe})/(z_{\rm harm}-z_{\rm safe})$. The lower anchor, $z_{\rm safe}\approx-89.68$, is the shared initial model's projection. The upper anchor, $z_{\rm harm}\approx-8.94$, is the median of the maximum projections reached by the four ordinary training trajectories. Both anchors come from completed ordinary runs and were fixed before feedback; neither uses judged answers or judge scores.

This normalization is the affine map $p(\theta)=\kappa a^\top b(\theta)+c$, where $\kappa=100/(z_{\rm harm}-z_{\rm safe})\approx1.24$ and $c=-\kappa z_{\rm safe}$. The multiplier converts projected changes into normalized points. Values above 100 indicate projections above the upper reference anchor.

The desired upper limit is $p^\star=30$. Steering activates at 28, ahead of that limit, and remains active until the readout falls to 25. A recovery target of 23 provides additional room below the release threshold. Target derivatives refresh every four updates; inactive updates omit attribution.

\paragraph{Prediction and information-guided allocation.}
The projected response $r_{t,j}=a^\top q_{t,j}$ predicts an example's effect along the control direction. Expanding $b(F_t(e^{\mathbf s}))$ at $\mathbf s=0$ predicts the correction to the measured ordinary endpoint, $p_{t+1}^{\rm ord}=p(F_t(\mathbf1))$:
\begin{equation}
\widehat p_{t+1}(\mathbf w)
=p_{t+1}^{\rm ord}+\kappa\sum_jr_{t,j}\ln w_j.
\label{eq:control-predictor}
\end{equation}
The predictor estimates the achievable correction; signed information sets the relative cost of weight changes in Equ.~\ref{eq:control-objective}. We use $S_{t,j}=\operatorname{sign}(r_{t,j})\tfrac12\log_2(1+\BGU_{t,j})$ in bits, while weight logarithms are natural. The vectors $\mathbf S_t$ and $\mathbf r_t$ collect the signed information and projected responses over the batch.

To define the penalty, we compare each example's share of total response magnitude with its share of total information. Responses at or below the numerical-zero threshold $\epsilon_0=10^{-14}$ are excluded from these shares; write $L_t=\{j:|r_{t,j}|>\epsilon_0\}$ for the remaining examples. Their penalties are
\[
m_{t,j}(\mathbf S_t,\mathbf r_t)=
\operatorname{clip}_{[1/4,4]}\!\left(
\frac{|r_{t,j}|/\sum_{k\in L_t}|r_{t,k}|}
{|S_{t,j}|/\sum_{k\in L_t}|S_{t,k}|}
\right),\quad j\in L_t,
\]
with $m_{t,j}=1$ outside $L_t$. A larger information share relative to response share makes a weight change less costly in the objective; clipping prevents that preference from becoming arbitrarily strong. The response controller uses the same objective with every $m_{t,j}=1$.

Both controllers restrict the weights to the same feasible set $\mathcal W_t$:
\[
\sum_j w_j=B_t,\qquad
w_j\in
\begin{cases}
[e^{-1},1],&r_{t,j}>\epsilon_0,\\
[1,e],&r_{t,j}<-\epsilon_0,\\
\{1\},&|r_{t,j}|\le\epsilon_0.
\end{cases}
\]
The first constraint preserves total batch weight. The bounds then allow strengthening examples to be downweighted and their weight to be redistributed to weakening examples, while numerical-zero responses retain unit weight. Together with the disturbance penalty, these constraints limit how far feedback changes the training update.

\paragraph{Buffered target and numerical solution.}
The target must account for how much correction the current batch permits. The shared deterministic maximum-correction routine computes this available correction, $\mathcal A_t$, in normalized behavioral points. For a threshold $u$, the correction needed to reach just below it is $H_u=\max\{p_{t+1}^{\rm ord}-(u-\delta),0\}$, with safety margin $\delta=0.2\kappa$. Thus the upper limit can be maintained when $\mathcal A_t\ge H_{30}$. In that case, the requested correction and target are
\[
D_t=\min\{H_{23},\max(H_{30},0.8\mathcal A_t)\},
\qquad p_t^{\rm tar}=p_{t+1}^{\rm ord}-D_t.
\]
The requested correction is at least what is needed for the upper limit and at most what reaches the recovery level. Between those bounds, it uses 80\% of the available correction. If the upper limit cannot be maintained, both controllers instead apply the maximum available correction.

For a feasible target, SLSQP solves Equ.~\ref{eq:control-objective}, starting from a feasible response-controller solution. We use tolerance $10^{-13}$ and at most 2,000 iterations. The numerical solve does not establish global optimality; if the information-guided solution is invalid, the controller retains the feasible response solution.

\paragraph{Overlapping contributions.}
Discounting an example's information under batch interference does not remove that example's effect from the predictor. For a batch of $k$ identical response vectors $q$, with isotropic resolution $\lambda I$ and the same log-weight change $s$, the predicted movement is $ksq$, although each example has BGU $\|q\|^2/(\lambda+(k-1)\|q\|^2)$. The controller therefore retains the full predicted effect of a group whose contributions overlap. Information guides the allocation of weight changes, while the response sum determines whether those changes achieve the requested correction.

\paragraph{Comparisons and evaluation.}
Response and signed-information feedback share the predictor, weight constraints, and activation/release policy. Their active steps can still differ because earlier interventions change the states reached later. The matched-random arm instead permutes the signed-information controller's weights among examples at each update, preserving its activity schedule and total batch weight. This comparison tests whether it matters which examples receive the weights. Both targeted controllers produce similar suppression (Table~\ref{tab:control-active}).

\begin{table}[!t]
\caption{\textbf{Both targeted controllers suppress internal behavior.} One Early trajectory per arm, with 564 observed boundaries through update 563. Exposure integrates $\max(p_t-30,0)$ by the trapezoidal rule, in persona-point updates. Random uses the signed-information donor's weights. These are internal readouts, not judged-answer scores.}
\label{tab:control-active}
\centering\small
\begin{tabular}{@{}lrrr@{}}
\toprule
Training & Final persona & Boundaries above 30 & Exposure above 30 \\
\midrule
Ordinary training & 43.03 & 555 & 22,923.5 \\
Response feedback & 10.29 & 154 & 6,100.1 \\
Signed-information feedback & 10.63 & 160 & 6,107.1 \\
Matched random & 35.67 & 555 & 21,404.6 \\
\bottomrule
\end{tabular}
\end{table}

Signed-information feedback is active for 364 of 563 updates; the figure shades each active interval $(t-1,t]$. Its separation from matched random supports assigning weights according to the examples' behavioral contributions.

The answer panel evaluates generated behavior in a separate feedback run, using mean judge ratings. A rating of at least 50 on the 0--100 scale counts as harmful. Ordinary training is evaluated at nine checkpoints, and feedback and random weights at $0,70,141,211,563$. At step 211, mean coherence is $85.53$ for ordinary training, $96.75$ for feedback, and $85.68$ for random weights. The panel's shading follows the internal-control trajectory described above, not the separate judged run.

\paragraph{Final-checkpoint utility.}
To check whether behavioral control reduces general task performance, we evaluate the ordinary and signed-information models from the internal-readout comparison at update 563. Both answer all 14,042 five-shot MMLU test questions under identical settings. Accuracy remains nearly unchanged: $73.81\%$ for ordinary training and $73.79\%$ for BGU feedback. The paired 95\% interval for BGU minus ordinary is $[-0.19,+0.16]$ percentage points, consistent with no observed control--utility tradeoff in this evaluation.

\section{Measured Cost of BS-Ghost}
\label{app:phase3b-systems}

Table~\ref{tab:systems-main} measures BS-Ghost's added cost in one matched pair on a Tesla V100-SXM2-32GB, using the fidelity configuration (Appendix~\ref{app:concept-influence}). The 1,000-example pass takes 63 updates, including a final batch of eight.

Rank-32 RS-LoRA uses scale 64 on all seven attention/MLP projection types, with FP16 base computation and FP32 adapters. Eight microbatches of two form each full batch. AdamW8bit uses $\beta=(0.9,0.999)$, $\epsilon=10^{-8}$, weight decay $0.01$, global clipping at one, and a linear schedule. BS-Ghost refreshes derivatives for the 15 layer-20 target coordinates at updates $1,5,\ldots,61$ and reuses them between refreshes.

\paragraph{Timing scope.}
Both runs start from identical training and random-number states after equal warmup. Timing includes training, attribution, score saving, and final synchronization, excluding model loading and common warmup. Ordinary training takes 331.90\,s and BS-Ghost takes 358.51\,s: the added cost is 26.61\,s, or 8.02\%. Every example receives a score, and the training states and score-based selections agree with their references.

\paragraph{Memory and packing.}
Peak allocated memory is 16.9\,GiB for ordinary training and 29.7\,GiB with BS-Ghost; peak reserved memory is 17.3 and 31.3\,GiB. Both runs use expandable memory segments. BS-Ghost reduces contraction overhead by packing compatible blocks (Appendix~\ref{app:packing-memory}), with the group sizes below for target contractions. Other groups contain at most 16 blocks and still contribute to global clipping. Packing changes contraction execution while preserving the training batch.
\begin{center}
\small
\begin{tabular}{@{}lrrrrr@{}}
\toprule
Block shape & $32\times3584$ & $3584\times32$ & $512\times32$ & $18944\times32$ & $32\times18944$ \\
\midrule
Blocks per group & 120 & 60 & 40 & 40 & 20 \\
\bottomrule
\end{tabular}
\end{center}

\paragraph{Published timing references.}
Published scoring times in Table~\ref{tab:systems-main} are not hardware-matched. Concept Influence does not specify the timing hardware or configuration details needed to compare its timing and fidelity setups; query costs exclude inverse-Hessian preparation \citep{kowal2026conceptinfluenceleveraginginterpretability}.

\end{document}